\documentclass[onecolumn]{IEEEtran}
\usepackage{amsmath}
\usepackage{amssymb}
\usepackage{mathrsfs}
\usepackage{amsfonts}
\usepackage{multirow}
\usepackage{cases}
\usepackage{epsfig}
\usepackage{graphicx}
\usepackage{epstopdf}
\usepackage{color}
\usepackage{algorithm}
\usepackage{booktabs}
\usepackage{listings}
\usepackage{boxedminipage}
\usepackage{epsfig,latexsym}
\usepackage{algorithmic}
\usepackage{stmaryrd}
\usepackage{psfrag}
\usepackage{float}
\usepackage{cite}
\usepackage[justification=centering,font=footnotesize]{caption} 
\usepackage{pgfplots}
\usepackage[font=footnotesize]{subcaption}
\usepackage{tikz}
\usepackage{tkz-tab}
\usepackage{relsize}

\newtheorem{proposition}{Proposition}
\newtheorem{theorem}{Theorem}
\newtheorem{assumption}{Assumption}

\newtheorem{remark}{Remark}
\newtheorem{subroutine}{Subroutine}
\newtheorem{corollary}{Corollary}

\newenvironment{mytabular}{\bgroup\tiny\tabular}{\endtabular\egroup}
\newenvironment{mytabular1}{\bgroup\scriptsize\tabular}{\endtabular\egroup}

 \newlength\figureheight 
     \newlength\figurewidth 

\begin{document} 
\title{Higher order Matching Pursuit for Low Rank Tensor Learning 
}

\author{ 
Yuning~Yang, Siamak~Mehrkanoon  
     and Johan~A.K.~Suykens,~\IEEEmembership{Fellow~IEEE}%
\thanks{Y. Yang, S. Mehrkanoon and J.A.K. Suykens are
with the Department of Electrical Engineering, STADIUS,
KU Leuven,  B-3001 Leuven, Belgium (email: yuning.yang, siamak.mehrkanoon,  johan.suykens@esat.kuleuven.be).
}
}
\maketitle

\begin{abstract}
Low rank tensor learning, such as tensor completion and multilinear multitask learning, has received much attention in recent years. In this paper, we propose higher order matching pursuit for low rank tensor learning  problems with a convex or a nonconvex cost function, which is a generalization of  the matching pursuit type methods. At each iteration, the main cost of the proposed methods is only to compute a rank-one tensor, which can be done efficiently, making the proposed methods scalable to large scale problems. Moreover, storing the resulting rank-one tensors is of low storage requirement, which can help to break the curse of dimensionality.  
The linear convergence rate of the proposed methods is established in various circumstances.   Along with the main methods, we also provide a  method of low computational complexity for approximately computing the rank-one tensors, with provable approximation ratio, which helps to improve the efficiency of the main methods and to analyze the convergence rate. Experimental results on synthetic as well as real datasets verify the efficiency and  effectiveness of the proposed methods. \\

\hspace{2mm}\vspace{3mm}
\end{abstract}

\section{Introduction}
Tensors, appearing as the higher order generalization of vectors and matrices, make it possible to represent data that have
intrinsically many dimensions, and give a better understanding
of the relationship behind the information from a higher order
perspective. In many machine learning problems such as tensor completion \cite{liu2013tensor,gandy2011tensor,signoretto2013learning,tomioka2013convex}, multilinear multitask learning (MLMTL) \cite{romera2013multilinear,wimalawarne2014multitask,signoretto2014graph} and tensor regression \cite{zhou2013tensor}, one often aims at learning a  tensor that has low rankness. For example, in tensor completion, the goal is to learn a low rank tensor provided that only partial observations are available. In the context of MLMTL, to allow for common information shared between tasks to pursuit better generalization, by learning several tasks simultaneously, where each task is indexed by more than two indices,   all the tasks can be represented by a tensor assumed to lie in a low dimensional spaces. In tensor regression, to better understand the information behind  high dimensionality data, the weight vector is represented by a low rank tensor. These applications give rise to low rank tensor learning.

Commonly speaking, to learn a low rank tensor, tensor learning minimizes a real-valued cost function $F:\mathbb T\rightarrow \mathbb R$ subject to some constraints or with regularizations to encourage the low rank property of the learned  tensor. Here $\mathbb T:=\mathbb R^{n_1\times\cdots\times n_N}$ denotes an $N$-th order tensor space, and $F(\cdot)$ is a continuous function. A widely used regularization is the sum of mode-$d$ matrix nuclear norms   \cite{liu2013tensor,gandy2011tensor,signoretto2013learning,tomioka2010estimation,romera2013multilinear,wimalawarne2014multitask,yang2012fixed}, which encourages the tensor to have low Tucker rank \cite{kolda2010tensor}. 
Some variations of the  nuclear norm, such as
the Schatten-$p$ norm \cite{signoretto2013learning,tomioka2013convex}, the latent norm and the scaled norm \cite{wimalawarne2014multitask},  as well as other variations    \cite{romera2013new,zhang2014hybrid}, have been studied. The advantage of the above approaches are their convexity, which enables them to be solved by many existing algorithms, while a main drawback is that all the approaches rely on solving singular value decompositions (SVD), which lacks   scalability. Another category of approaches decomposes the tensor  into several factors, and applies the alternating minimization rule to solve the resulting optimization problems, see, e.g., \cite{kolda2010tensor,romera2013multilinear,chen2014simultaneous}. This type of approaches avoids computing SVDs, but lacks   global convergence analysis.
Recently, tensor nuclear norm based algorithms are proposed in \cite{yu2014approximate,yang2015a}. This type of algorithms solves a simple and efficient subproblem at each iteration, which is scalable to large-scale problems. However they are specifically designed for convex $F(\cdot)$, and the stepsizes have some restrictions.

In this paper, motivated by the simple and efficient matching pursuit (MP) methods 
 for sparse approximation \cite{mallat1993matching,pati1993orthogonal,tropp2004greed}, signal recovery \cite{tropp2007signal}, matrix compoletion \cite{wang2014orthogonal} and greedy method for tensor approximation \cite{kolda2001orthogonal}, we propose higher order matching pursuit (HoMP)   methods for solving low rank tensor learning problems, either with a convex or nonconvex cost function.  At each iteration, the classical MP selects an ``atom'' from a given dictionary, and then updates the new trial based on the linear combination of the current trial and the selected atom, with suitably chosen weights, whereas in tensor learning setting,  the atom is a rank-one tensor, which has to be learned based on the gradient information of $F(\cdot)$.  Finding such a rank-one tensor reduces to computing a tensor spectral norm, or known as the tensor singular value problem \cite{Lim2005,lim2014blind}. Although solving such a subproblem  exactly is NP-hard in general \cite{hillar2013most}, approximation methods exist, and fortunately, as MP, HoMP allows to use an approximation solution. This feature makes HoMP particularly suitable for tensor learning. When choosing the weights,   if the cost function $F(\cdot)$ is associated with a least squares loss, then three strategies, which are in accordance with matching pursuit \cite{mallat1993matching}, economic orthogonal MP \cite{wang2014orthogonal}  (or relaxed MP, see, e.g., \cite{barron2008approximation}) and orthogonal matching pursuit \cite{pati1993orthogonal}, can be applied. We then generalize HoMP to the case that $F(\cdot)$ is nonconvex, where the weights are chosen by minimizing a  quadratic function which majorizes $F(\cdot)$. Along with the main HoMP methods, an efficient algorithm will be presented to approximately and efficiently solve the tensor singular value problem mentioned above, with a provable approximation ratio. This ratio is important in analyzing the convergence rate, which will be used extensively in Sect. \ref{sec:lc}. Besides the efficiency, another advantage of HoMP-type methods is its low storage requirement, as will be explained in Sect. \ref{sec:mp}.
 
 The convergence rate of HoMP is analyzed in two specific problems: tensor completion and MLMTL. Specifically, we first show that, if $F(\cdot)$ is associated with the least squares loss, then HoMP converges linearly for the two specific problems. We then generalize our analysis to a class of loss functions, which may be nonconvex and includes many robust losses as special cases. Interestingly, the linear convergence rate can still be established  in these cases. 

In a nutshell, our contribution is summarized as follows:
\begin{enumerate}
\item We propose efficient HoMP  methods for tensor learning, which are applicable for problems with   with convex or nonconvex cost functions;
\item We present an efficient method for selecting   rank-one tensors, with provable approximation ratio. The ratio is important in analyzing the convergence rate.
\item We establish linear convergence of the proposed HoMP  methods, either for problems with convex or nonconvex cost functions.
\end{enumerate}

     The rest of this paper is organized as follows. Sect. \ref{sec:pre}   gives preliminaries on tensors. Low rank tensor learning problems are formulated in Sect. \ref{sec:lrtl}, and specified by tensor completion and MLMTL. The HoMP-type methods, their related work, and an efficient algorithm for selecting rank-one tensors will be detailed in Sect. \ref{sec:mp}. Sect. \ref{sec:lc} is focused on analyzing the convergence rate. Numerical experiments will be conducted in Sect. \ref{sec:exp}. Finally, Sect. \ref{sec:con} draws some conclusions.

\subsection{Preliminaries on tensors}\label{sec:pre}
 Vectors are written as   $(\mathbf a,\mathbf b,\ldots)$, matrices
  correspond to   $(\mathbf A,\mathbf B,\ldots)$, and tensors are
  written as   $(\mathcal{A}, \mathcal{B},
  \cdots)$.   $\mathbb T :=\mathbb R^{n_1\times n_2\times \cdots\times n_N}$   denotes  an $N$-th order  tensor space.

   For two tensors $\mathcal A,\mathcal B\in\mathbb T$, their inner product is given by
      $\langle \mathcal A,\mathcal B\rangle = \sum^{n_1}_{i_1=1}\cdots\sum^{n_N}_{i_N=1}\mathcal A_{i_1\cdots i_N}\mathcal B_{i_1\cdots i_N}.$
      The Frobenius norm of $\mathcal A$ is defined by $\|\mathcal A\|_F = \langle\mathcal A,\mathcal A\rangle^{1/2}.$    
     The outer  product of vectors $\mathbf x_i\in\mathbb R^{n_i}$, $i=1,\ldots,N$ is denoted as $\mathbf x_1\otimes\cdots \otimes \mathbf x_N$ and is a rank-one tensor in $\mathbb T$ defined by
$(\mathbf x_1\otimes\cdots \otimes \mathbf x_N)_{i_1\cdots i_N} = \prod^N_{j=1}\mathbf x_{j,i_j}$. The mode-$d$ tensor-matrix multiplication of a tensor $\mathcal X\in\mathbb T$ with a matrix $ U\in\mathbb R^{J_d\times n_d}$ is a tensor of size $n_1\times \cdots\times n_{d-1}\times J_d\times n_{d+1}\times \cdots\times n_N$. 

\subsubsection{Tensor-matrix  mode-$d$ unfolding and mode-$(p,q)$ unfolding}  The mode-$d$ unfolding of tensor $\mathcal A$
is denoted as $\mathbf A_{(d)}$ by   arranging the
mode-$d$ fibers to be the columns of the resulting matrix. 
For an $N$-th order tensor $\mathcal A$, the mode-$(p,q)$ unfolding is to   choose the $(p,q)$ modes and merge them into the first mode (row) of the unfolding matrix, and the remaining $N- p-q$ modes are merged into the second mode (column). The unfolding is denoted as $\mathbf A_{[p,q;N\setminus\{p,q\}]}$ and can be simplified as $\mathbf A_{(p,q)}$ if necessary, where the semicolon specifies a new mode. We explain it by an example. For a $4$-th order tensor $\mathcal A$, its mode-$(1,2)$ unfolding is $\mathbf A_{[1,2;3,4]}$, defined by
 $(\mathbf A_{[1,2;3,4]})_{(i_1-1)n_2+i_2,(i_3-1)n_4+i_4} = \mathcal A_{i_1 i_2 i_3 i_4} $. 
 
\subsubsection{Tensor rank}  There are mainly two types of tensor rank, namely the CP-rank and the Tucker-rank \cite{kolda2010tensor}. The CP-rank is defined as the minimum   integer $R$ such that for a tensor $\mathcal X$, it can be factorized as a sum of $R$ rank-one tensors.   ${\rm rank_{CP} }(\cdot)$ will be used to denote the CP-rank in this paper. 
   The Tucker-rank of an $N$-th order  tensor $\mathcal X$  is an $N$ tuple,  whose $i$-th entry is   the rank of the unfolding matrix $  X_{(i)}$.

\subsubsection{Tensor singular value problem} For a tensor $\mathcal X\in\mathbb T$, its largest singular value   is defined as   \cite{Lim2005}
\begin{equation}\label{eq:singular}
\max_{\|\mathcal Y\|_F=1,{\rm rank_{CP}}(\mathcal Y)=1}\nolimits \left\langle \mathcal X, \mathcal  Y\right\rangle,
\end{equation}
which is equivalent to the tensor spectral norm $\|\mathcal X\|_2$ and is dual to the tensor spectral norm, see, e.g., \cite{lim2014blind}. Solving such a problem is NP-hard in general \cite{hillar2013most}.

\section{Low Rank Tensor Learning Formulation}\label{sec:lrtl}
As introduced in the introduction,  low rank tensor learning seeks a low rank tensor solution via minimizing a cost function. Mathematically, the model considered in this work is of the following general form
\begin{equation}\label{prob:lrtl}
\min_{\mathcal W }\nolimits F(\mathcal W)~~{\rm s.t.}~~ {\rm rank_{CP}}(\mathcal W)\leq K, 
\end{equation}
where $F:\mathbb T\rightarrow \mathbb R$ denotes the cost function  which is continuously differentiable,   either convex or nonconvex. The low CP-rank constraint encourages the learned tensor to have low rank structure. We specify the model \eqref{prob:lrtl} via two specific applications: tensor completion and MLMTL.

\subsection{Tensor completion} The goal of tensor completion is to infer a tensor (possibly low rank) from its partial observations. Mathematically, given a partially observed tensor $\mathcal B_{\Omega}$ where $\Omega$ denotes the index set of observed entries, the problem can be formulated as 
$$\min_{\mathcal W\in\mathbb T}\nolimits~r(\mathcal W) ~~{\rm s.t.}~ \mathcal W_{\Omega} = \mathcal B_{\Omega},$$
where $r(\cdot)$ is used to control the low rankness of the tensor, such as ${\rm rank_{CP}}(\cdot)$, and the sum of nuclear norms. In our setting, letting $F(\mathcal W):=\sum_{(i_1,\ldots,i_N)\in\Omega}\ell(\mathcal W_{i_1\cdots i_N}-\mathcal B_{i_1\cdots i_N})$ with a specific loss  $\ell(\cdot)$, we model the problem as follows\\
$
~~~~~~~~~~~~~~~~\min_{{\rm rank_{CP}}(\mathcal W)\leq K}\nolimits F(\mathcal W),
$\\
with $K$ being a positive integer to control the CP-rank of $\mathcal W$.  
\subsection{Multilinear multitask learning (MLMTL)} MLMTL learns many tasks simultaneously, where each task is indexed by more than two indices \cite{romera2013multilinear,wimalawarne2014multitask}. An example is to predict consumers' ratings for restaurants, where each rating contains several aspects. Then each task is indexed by consumer and aspect.  Since each task can be represented by a weight vector, all the tasks jointly yield a third-order tensor, see, e.g., \cite{romera2013multilinear}. In the following, we restrict ourselves to multilinear multitask regression. Specifically, 
We consider $T$   tasks, each of which is specified by a weight vector $\mathbf w^t\in\mathbb R^D$ that corresponds to a linear function $\langle \mathbf x,\mathbf w^t\rangle$, where $\mathbf x$ is an observed input. Provided that the associated observed output is $y$, we employ a specific loss $\ell(\langle \mathbf x,\mathbf w^t\rangle-y)$.  
  For each task $\mathbf w^t$, a finite set of training samples $\{ (\mathbf x^t_i,y^t_i) \}^{m_t}_{i=1}$ is available, and we aim at minimizing the empirical risk $F(\mathbf W)$ defined as
\begin{equation*}
F(\mathbf W) :=  \sum^T_{t=1}\nolimits {m_t^{-1}}\sum^{m_t}_{i=1}\nolimits\ell(\langle \mathbf x^t_i,\mathbf w^t\rangle-y^t_i),
\end{equation*}
 where 
$\mathbf W = [\mathbf w^1,\ldots,\mathbf w^T]\in\mathbb R^{D\times T}$
is the weight matrix.
For each task $t$, we assume that it is related to $N\geq 2$ indices, each of which varies from $1$ to $n_i$, $i=1,\ldots,N$. That is, the task $\mathbf w^t$ can be identified by the indices $(i_1,\ldots,i_N) \in [n_1]\times \cdots\times [n_N]$. In this case, we have $T=\prod^N_{i=1}n_i$. Correspondingly, the matrix $\mathbf W$ can be folded to an $(N+1)$-th order tensor $\mathcal W$, with size $D\times n_1\times\cdots\times n_N$, and $\mathbf W$ can be regarded as the mode-$1$ unfolding of $\mathcal W$. We also denote   $F(\mathcal W) =F(\mathbf W)$.
Assuming that the tasks share certain common structure, our model is defined as\\
$
~~~~~~~~~~~~~~~~\min_{{\rm rank_{CP}}(\mathcal W)\leq K}\nolimits F(\mathcal W),
$ 
 
Therefore, both of our models of tensor completion and MLMTL adopts the CP-rank to control the low rankness of the learned tensor, which is quite different from models based on nuclear norm regularizations \cite{liu2013tensor,gandy2011tensor,signoretto2013learning,tomioka2010estimation,romera2013multilinear,wimalawarne2014multitask}.
 
\section{Higher Order Matching Pursuit} \label{sec:mp}

Having presenting our low rank tensor learning problem \eqref{prob:lrtl}, the goal of this section is to introduce the Higher order Matching Pursuit (HoMP) methods  to solve it. HoMP methods are presented in   Algorithm \ref{alg:mp}.

{\renewcommand\baselinestretch{0.1}\selectfont
 \begin{algorithm*}\caption{Higher order Matching Pursuit (HoMP) for low rank tensor learning } \label{alg:mp}
 \begin{algorithmic}[H]
 \STATE{\textbf{Input:} $\mathcal W^{(0)}=0$; $\overline \alpha_0=0$; $K\geq 1$.}
 \STATE{\textbf{Output:} the resulting tensor   $ \mathcal W^{(K)}. $}
 \FOR{$k=1$ \TO $K$ }
 \STATE{$\bullet$ Select a normalized rank-one tensor $\mathcal S^{(k)}$: 
 \begin{equation}\label{eq:mp:sigular}
\langle \nabla F(\mathcal W^{(k)}), \mathcal S^{(k)} \rangle \geq \beta \max_{\|\mathcal S\|_F=1,{\rm rank_{CP} (\mathcal S)=1}}\langle \nabla F(\mathcal W^{(k)}), \mathcal S \rangle~~(0<\beta\leq 1)
 \end{equation}
 }
 \STATE{$\bullet$ Update \\
 ~~~~ 1) $\mathcal W^{(k+1)} = \mathcal W^{(k)} + \overline{\alpha}\mathcal S^{(k)}, ~\overline{\alpha} = \arg\min_{\alpha}F(\mathcal W^{(k)} + \alpha\mathcal S^{(k)})$ ~~~~~~~~~~~~~~~~~~~~~~~~~~~(HoMP-LS)\\
 ~~~~ 2) $\mathcal W^{(k+1)} = \overline{\alpha_1}\mathcal W^{(k)} + \overline{\alpha_{2}}\mathcal S^{(k)}, ~(\overline{\alpha_1},\overline{\alpha_2})=\arg\min_{(\alpha_1,\alpha_2)}F(\alpha_1\mathcal W^{(k)} + \alpha_2\mathcal S^{(k)})$ ~~~~(HoRMP-LS)\\
 ~~~~ 3) $\mathcal W^{(k+1)} = \sum^k_{i=0}\overline{\alpha_i}\mathcal S^{(i)},~ \boldsymbol{\overline\alpha}=(\overline{\alpha_0},\ldots,\overline{\alpha_k})^{\top} = \arg\min_{\alpha\in\mathbb R^{k+1}}F(\sum^k_{i=0}\alpha_i\mathcal S^{(i)}) $ ~~~(HoOMP-LS)\\
 ~~~~ 4) $\mathcal W^{(k+1)} = \mathcal W^{(k)} + \overline{\alpha}\mathcal S^{(k)}, ~\overline{\alpha} = -\langle\nabla F(\mathcal W^{(k)}),\mathcal S^{(k)}\rangle/L$ ~~~~~~~~~~~~~~(HoMP-G)~~~~($L$~{\rm is~a~Lipschitz~constant})
  }
  \ENDFOR
 \end{algorithmic}
 \end{algorithm*}
 \par}

We describe the method in more details.
 Given a cost function $F(\cdot)$ of low rank tensor learning, with initial guess $\mathcal W^{(0)}$ being the zero tensor, at each iteration, HoMP can be divided into two steps: the selection step and the updating step. The selection step chooses certain atom, which is a rank-one tensor $\mathcal S^{(k)}$ by solving the tensor singular value problem \eqref{eq:singular} approximately with an approximation ratio $\beta$, where $\mathcal X=\nabla F(\mathcal W^{(k)})$, as shown in \eqref{eq:mp:sigular}. The approximation ratio is important in convergence rate analysis, as will be shown in Sect. \ref{sec:lc}.   The updating step adaptively computes the weights and updates the new trial. This step has two   cases:  
 
 1) If $F(\cdot)$ is associated with a least squares loss, and  $F(\cdot)$ with respect to weights $\alpha$ is quadratic,  then three strategies can be considered:   higher order MP with least squares loss (HoMP-LS),   higher order relaxed MP with least squares loss (HoRMP-LS) and higher order orthogonal MP with least squares loss (HoOMP-LS), as shown in Algorithm \ref{alg:mp}, where all the weights can be computed by solving certain least squares problems.    These strategies are respectively in accordance with  MP \cite{mallat1993matching}, economic orthogonal MP \cite{wang2014orthogonal}  (or relaxed MP, see, e.g., \cite{barron2008approximation}) and orthogonal MP \cite{pati1993orthogonal}, and generalize them to tensor learning setting. 
 
 2) If  $F(\cdot)$ is associated with a general loss which is possibly nonconvex, while the gradient $\nabla F(\cdot)$ is Lipschitz continuous with Lipschitz constant  $L$, then the strategy higher order MP with a general loss (HoMP-G) can be applied. In fact, the weight is computed by minimizing a quadratic function that majorizes $F(\cdot)$ at $\mathcal W^{(k+1)}$.
 
If Algorithm \ref{alg:mp} stops within $K$ iterations, then it generates a feasible solution to   \eqref{prob:lrtl}. Comparing between HoMP, HoRMP and HoOMP, one can see that HoOMP may obtain a better new trial, as it updates the weights most greedily, while HoMP updates the weights least greedily. However, computing the weights for HoOMP may be time consuming, as it may require  to solve a linear equations system. HoRMP, which considers the linear combination between the current trial and the new atom, can be regarded as a trade-off  between HoMP and HoOMP.
 
The main computational cost of HoMP-type methods is the selection step \eqref{eq:mp:sigular}. Comparing with those methods that require to solve SVD, solving \eqref{eq:mp:sigular} will be more efficient, as will be detailed in subsection \ref{sec:alg}. 

Another advantage of HoMP-type methods is the low storage requirement. Suppose we work in a tensor space $\mathbb T$ and the methods stop within $K$ iterations with $K$ not too large; since   the learned tensor is a combination of some rank-one tensors, and each rank-one tensor can be  represented by the outer product of $N$ vectors, the whole tensor can be stored by using $\sum^N_{i=1}n_i\cdot K$ storage only, against using $\prod^N_{i=1}n_i$ to store the whole tensor. This can help to break the curse of dimensionality.

\subsection{Related work}\label{sec:related}
As mentioned earlier, HoMP-type methods are motivated by MP-type methods
 for sparse approximation \cite{mallat1993matching,pati1993orthogonal,tropp2004greed}, signal recovery \cite{tropp2007signal} and matrix completion \cite{wang2014orthogonal}. The classical MPs iteratively select atoms from a redundant dictionary, one at a time, and then use their certain linear combination to approximate a given signal. Recently, \cite{wang2014orthogonal} generalizes MPs to   matrix completion, where the cost function is least squares based, and proves the linear convergence of their methods. Our work generalizes \cite{wang2014orthogonal} in the following senses: 1) we generalize MPs to tensor learning problems; 2) our methods can be adapted to problems with  a general loss; 3) the way of selecting atoms in tensor setting is more challenge than in matrix cases; 4) we establish linear convergence rate for a wide range of loss functions, which may possibly be nonconvex.

 Another issue  related to HoMPs is the approach   of successive rank-one approximation to tensors (SR1A), see,   \cite[Sect. 3.3]{grasedyck2013literature} and the references therein. In general, consider   a linear system  $\mathcal A(\mathcal W)\approx \mathbf b$ where $\mathcal A$ is a linear operator, $\mathbf b$ is a vector and $\mathcal W$ is a low rank tensor to be determined. SR1A updates the new trial as $\mathcal W^{(k+1)} = \mathcal W^{(k)} + \mathcal S^{(k)}$ where $\mathcal S^{(k )}$ is a rank-one tensor which minimizes some convex cost function $E(\mathcal A(\mathcal X) - \mathbf b)$, see, e.g., \cite{figueroa2012greedy,ammar2010convergence}. Comparing with SR1A, HoMPs are more general in terms of the problems to be solved, the cost function $F(\cdot)$ to be minimized, and the strategies of choosing the weights. Moreover, finding the rank-one term $\mathcal S$ in SR1A is either intractable or needs an alternating minimization strategy \cite{kolda2001orthogonal} without explicit approximation ratio, while we allow an approximation solution \eqref{eq:mp:sigular}.
 
 HoMPs are also closely related to conditional gradient (CG) methods for tensor learning \cite{yu2014approximate,yang2015a}. CG, also  known as the Frank-Wolfe method \cite{frank1956algorithm}, is a classical method for constrained convex optimization problems, and regains attention in recent years, see, e.g., \cite{jaggi2013revisiting}. At each iteration, CG also computes an atom. Then, different from MPs, CG performs a convex combination of the selected atom and the current trial, which restricts  the new trial to still lie in the convex constraint. This difference leads to significant differences to the models behind MPs and CG, where MPs minimize the cost function constrained by a ``hard'' constraint, while CG minimizes the cost function constrained by its ``soft'' counterpart, such as $L_0$ norm versus $L_1$ norm, matrix/tensor rank versus matrix/tensor nuclear norm.

\subsection{Efficiently computing the selection step \eqref{eq:mp:sigular}}\label{sec:alg}
      
Since solving \eqref{eq:singular} exactly is NP-hard in general \cite{hillar2013most}, the goal of this subsection is to present a    method to find an approximation solution of \eqref{eq:mp:sigular}  with low computational complexity, and to explicitly derive the approximation ratio. 
 In the literature,
several   approximation methods have been proposed, e.g., the power-type methods  \cite{Lmv2000,chlz2012,zlq2012}, the Newton method   \cite{zhang2001rank}, and others \cite{zqy2012,hlz2010,So2010,hjlz2012,yang2013rankone}.   However, these methods   are not very efficient in our setting.     

For a $2d$-th order tensor $\mathcal A$, our method is defined recursively as follows, where the output is a set of $2d$ normalized vectors, whose outer product yields the approximation solution.
\begin{center}
 \begin{boxedminipage}{.5\textwidth}
\begin{subroutine}  {\bf $ ( \mathbf x_1,\ldots,\mathbf x_{2d})=$ ApproxSpectral2d($\mathcal A$)} \label{sub:2}
\begin{enumerate}
\item If the order   is $2$, return  the   singular vector pair  $(\mathbf x_{1},\mathbf x_{2})$   corresponding to the leading singular value of $\mathcal A$;
\item Compute  the (inexact)  singular vector pair  $(\mathbf x_{[1,2]},\mathbf x_{[3,\ldots,2d]})$   corresponding to the leading singular value of matrix $\mathbf A_{[1,2;3,\ldots,2d]}$. 

\item Fold the vector  $\mathbf x_{[1,2]}$    to matrix $ \mathbf X_{[1;2]}$ and compute the leading singular vector pair $\mathbf x_{[1]}$ and $\mathbf x_{[2]}$;    

\item  Denote  the   $(2d-2)$-th order tensor $\mathcal Y := \mathcal A\times_1\mathbf x_1^{\top}\times_2\mathbf x_{2}^{\top}$;   
 
~~compute $ (\mathbf x_{3},\ldots,\mathbf x_{2{d}}) = $ {\bf ApproxSpectral2d}($\mathcal Y$).
\item Return $(\mathbf x_1,\ldots,\mathbf x_{2d})$.
\end{enumerate}
\end{subroutine}
 \end{boxedminipage}
\end{center}

  At each recursion of Subroutine \ref{sub:2}, the dominant   cost  is   Step 2), i.e., to compute the leading singular value of a matrix of size $n^2\times n^{(2d-2k)}$ at the $k$-th recursion, with $1\leq k\leq (d-1)$. Although the computational complexity of only computing the leading singular value is less than doing the full SVD, when the size goes high, this procedure still takes much time. Empirically we find that there is no need to compute the singular value precisely, and running   a few power iterations is acceptable. That is, we can compute Step 2) inexactly. Suppose a few power iterations have been performed in Step 2), and we obtain   a normalized vector pair $(\mathbf x_{[1,2]}, \mathbf x_{[3,\ldots,2d]})$ such that $\mathbf A_{(1,2)}\mathbf x_{[3,\ldots,2d]} = \alpha_{(d)} \|\mathbf A_{(1,2)}\|_2 \mathbf x_{[1,2]}$ where $0<\alpha_{(d)}\leq 1$, and $\mathbf A_{(1,2)}$ is short for $\mathbf A_{[1,2;3,\ldots,2d]}$. Then we can establish the following lower bound. For simplicity, we may assume $n_1=n_2=\cdots = n_{2 d}=n$.
\begin{proposition}  Let the order of the tensor be $2d$. 
Suppose at the $k$-th recursion, the vectors  obtained in Step 2) are normalized and satisfy
\begin{equation}\label{eq:prob:bound2:1}
\mathbf A_{(1,2)}\mathbf x_{[3,\ldots,2d-2k+2]} = \alpha_{(k)} \|\mathbf A_{(1,2)}\|_2 \mathbf x_{[1,2]},
\end{equation}
where $0<c\leq\alpha_{(k)}\leq 1$, $1\leq k\leq d-1$ and $c$ is a constant. Then there holds
\begin{equation}\label{eq:bound2}
\left\langle \mathcal A,\mathbf x_{1}\otimes \cdots\otimes \mathbf x_{2 d}\right\rangle\geq \frac{\prod^{d-1}_{k=1}\alpha_{(k)}\|\mathbf A_{(1,2)}\|_2}{n^{\max\{0,3d/2-2\}}}\geq  \frac{\prod^{d-1}_{k=1}\alpha_{(k)}\|\mathcal A\|_2}{n^{\max\{0,3d/2-2\}}}.
\end{equation}
\end{proposition}
 
\begin{IEEEproof}
We denote $v = \left\langle \mathcal A,\mathbf x_{1}\otimes\cdots\otimes \mathbf x_{2d}\right\rangle$, tensor  $\mathcal M_1:= \mathcal A\times_1\mathbf x_1^{\top}\times_2 \mathbf x^{\top}_{2}$ and matrix $\mathbf  M_2:= \langle\mathcal A, \cdot\otimes\cdot \otimes \mathcal X_{[3; \cdots;2 {d}]} \rangle$. Here $\mathcal X_{[3; \cdots;2 {d}]}$ is a $(2d-2)$-th order tensor folded by the  vector $\mathbf x_{[3,\ldots,2 d]}$ generated in Step 2) of Subroutine \ref{sub:2}, either exactly or inexactly, and  the $(i,j)$-th entry of matrix $\mathbf M_2$ is given by the inner product of $\mathcal A(i,j,:,:,\ldots,:)$ and $\mathcal X_{[3; \cdots;2 {d}]}$. For ease of notation, we also denote $\mathbf M_1$ the mode-$(1,2)$ unfolding of $\mathcal M_1$.  We use the induction method on  $d$. When $d=1$, \eqref{eq:bound2} holds. Suppose \eqref{eq:bound2} holds when $d=l\geq 2$. When $d=l+1$, there holds
\begin{eqnarray*}
v &=& \langle \mathcal M_1, \mathbf x_{3}\otimes\cdots\otimes \mathbf x_{2 {(l+1)}}\rangle \\
&\geq& \frac{\prod^{d-1}_{k=2} \alpha_{(k)}\| \mathbf M_{1 }\|_2}{n^{3/2l-2}}~({\rm from~ Step~ 4)~ and~ the~ induction})\\
&\geq& \frac{\prod^{d-1}_{k=2} \alpha_{(k)}\| \mathbf M_1\|_F}{n^{3/2l-2}\cdot n} 
 =  \max_{\|\mathcal X\|_F=1}\frac{\prod^{d-1}_{k=2} \alpha_{(k)}\langle \mathcal M_1,\mathcal X \rangle}{n^{3/2l-1 } }\\
 &\geq& \frac{\prod^{d-1}_{k=2} \alpha_{(k)} \langle \mathcal M_1,\mathcal X_{[3;\cdots;2 {(l+1)}]}\rangle}{n^{3/2l-1}}\\
 & =& \frac{ \prod^{d-1}_{k=2} \alpha_{(k)}  \langle\mathbf M_2,\mathbf x_1\otimes \mathbf x_{2}\rangle}{n^{3/2l-1}}\\ 
 &=& \frac{ \prod^{d-1}_{k=1} \alpha_{(k)} \|\mathbf A_{(1,2)}\|_2\langle \mathbf x_1^{\top}\mathbf X_{[1;2]} \mathbf x_{2}\rangle} {n^{3/2l-1}}~~~~~~~~~({\rm from~\eqref{eq:prob:bound2:1}})\\
 &\geq& \frac{\prod^{d-1}_{k=1} \alpha_{(k)}\|\mathbf A_{(1,2)}\|_2}{n^{3/2{l}-1/2}} \geq \frac{\prod^{d-1}_{k=1} \alpha_{(k)}\|\mathcal A\|_2}{n^{3/2{(l+1)}-2}},
\end{eqnarray*}
where the second and the last inequalities follows from the relationship between matrix spectral norm and Frobenius norm. Therefore, \eqref{eq:bound2} has been proved.
\end{IEEEproof}

 After applying Subroutine \ref{sub:2}, we can perform the following block coordinate updating subroutine \cite{Lmv2000} to further improve the solution quality. That is to say, we can get a larger value $\langle \mathcal A, \mathbf {\tilde x}_1,\ldots,\mathbf {\tilde x}_{2d}\rangle  $ than $\left\langle \mathcal A,\mathbf x_{1}\otimes \cdots\otimes \mathbf x_{2 d}\right\rangle$ obtain in \eqref{eq:bound2}.

\begin{center}
 \begin{boxedminipage}{.5\textwidth}
\begin{subroutine}  {\bf $ ( \mathbf {\tilde x}_1,\ldots,\mathbf {\tilde x}_{2d})=$ BCU($\mathcal A,\mathbf x_1,\ldots,\mathbf x_{2d}$)} \label{sub:3}

{\bf for} $i=1,\ldots,d$\\
$~~~~$Compute the singular vector pair $(\mathbf x_{2i-1},\mathbf x_{2i})$ corresponding to the leading singular value of the  matrix $\mathcal A\times_1\mathbf{\tilde x}_1\times_2\cdots\times_{2i-2}\mathbf{\tilde x}_{2i-2}\times_{2i+1}\mathbf    x_{2i+1} \times_{2i+2}\cdots\times_{2d}\mathbf x_{2d}$.\\
{\bf end for}

\end{subroutine}
 \end{boxedminipage}
\end{center}

Of course, Subroutine \ref{sub:3} can be applied several times after Subroutine \ref{sub:2} to get a better solution. However, considering the trade-off between the computational cost and the solution quality,   performing   it a few times is enough to get an acceptable solution. 

We discuss the computational complexity. At the $k$-th recursion of Subroutine \ref{sub:2}, the complexity is $O(n^{2(d-k)}) + O(n^2)$; Subroutine \ref{sub:3} is $O(n^{2d})+O(n^2)$. Thus the total complexity of Subroutine \ref{sub:2} together with Subroutine \ref{sub:3} is $\sum^d_{i=1}O(n^{2i})$, which is slightly worse than the power method for computing the leading singular value of a matrix of the same size of the tensor considered here. Therefore,
comparing with methods based on SVD, e.g., \cite{liu2013tensor,gandy2011tensor}, HoMPs   may be more efficient: for $N$-th order tensor space,  at each iteration, methods based on SVD require to perform $N$ SVD, with complexity at least $O(N n^{N+1})$, which is higher than ours.


 Finally, we note that tensors whose order  is odd   can always   be   treated as   a tensor of even order, e.g., a tensor $\mathcal A\in\mathbb R^{n_1\times n_2\times n_3}$ can be seen as in the space $\mathbb R^{1\times n_1\times n_2\times n_3}$. 

\section{Convergence rate analysis}\label{sec:lc}
In this section, we will establish the linear convergence rate of Algorithm \ref{alg:mp} for tensor completion and MLMTL, either with a convex cost function with least squares loss, or with a possibly nonconvex cost function. Tensors in this section are assumed to be of order $N$, $\mathcal W\in\mathbb T$. A key property for the analysis is inequality \eqref{eq:bound2}. To fit into the language of Algorithm \ref{alg:mp}, we write it as
\begin{equation}\label{eq:key_inequality}
\langle \nabla F(\mathcal W^{(k)}), \mathcal S^{(k)}\rangle \geq \rho \|\nabla F(\mathcal W^{(k)})_{(1,2)}\|_2,
\end{equation}
where $0<\rho\leq 1$ is a ratio depending on the size  of the tensor,  as that derived in \eqref{eq:bound2}, and $\nabla F(\mathcal W^{(k)})_{(1,2)}$ is the mode-$(1,2)$ unfolding of $\nabla F(\mathcal W^{(k)})$.

This section is organized as follows: in subsection \ref{sec:lc_ls} we prove the linear convergence for tensor completion and MLMTL when $F(\cdot)$ is associated with a least squares loss; in subsection \ref{sec:lc_non_ls} we extend the linear convergence results to $F(\cdot)$ with a possibly nonconvex loss.

 \subsection{Least squares} \label{sec:lc_ls}
 \subsubsection{Tensor completion}
 In the least squares sense, the cost function of tensor completion can be written as
 $$F(\mathcal W) = {1/2}\|\mathcal W_{\Omega} - \mathcal B_{\Omega}\|_F^2.$$
 The following theorem will establish the linear convergence rate of HoMP-LS, HoRMP-LS and HoOMP-LS uniformly in terms of the objective value $F(\mathcal W)$.
 
 \begin{theorem}[Linear convergence rate for tensor completion]\label{th:lc_tc_ls}
   Denote $\mathcal R^{(k)}:=    \mathcal W^{(k)}_{\Omega}-\mathcal B_{\Omega}$, and let $\mathcal S^{(k)}$ be generated by Subroutine \ref{sub:2} with input $\nabla F(\mathcal W^{(k)})$, and \eqref{eq:key_inequality} holds.  If $\{\mathcal W^{(k)} \}$ is generated by HoMP-LS, HoRMP-LS and HoOMP-LS,  then there holds
$$F(\mathcal W^{(k+1)})\leq \left( 1- \frac{\rho^2}{{n_1n_2}}\right)F(\mathcal W^{(k)}).$$
 \end{theorem}
\begin{IEEEproof}
We first consider HoOMP-LS. For clarity we denote the weights $\boldsymbol{\overline{\alpha}}$ at the $k$-th step as $\boldsymbol{\overline{\alpha}}^k = (\overline{\alpha_0^k},\ldots,\overline{\alpha_k^k})$. For convenience we denote $\|\mathcal X_{\Omega}\|_F = \|\mathcal X\|_{\Omega}$. 
By the definition of $\mathcal R^{(k)}$ and $\mathcal W^{(k+1)}$, there holds
\begin{eqnarray*}
2F(\mathcal W^{(k+1)})=\|\mathcal R^{(k+1)}\|_F^2  &=& \|     \mathcal W^{(k+1)}-\mathcal B \|_{\Omega}^2  = \min_{\boldsymbol{\alpha}}\|    \sum^k_{i=0} {\alpha_i }\mathcal S^{(i)}-\mathcal B\|_{\Omega}^2\\
&\leq&  \min_{\alpha}\|   \sum^{k-1}_{i=0}\overline{\alpha^{k-1}_i}\mathcal S^{(i)} + \alpha \mathcal S^{(k)}-\mathcal B \|_{\Omega}^2\\
&=& \min_{\alpha}\|  \mathcal W^{(k)} -\mathcal B + \alpha \mathcal S^{(k)}\|_{\Omega}^2 = \min_{\alpha}\|\mathcal R^{(k)} + \alpha\mathcal S^{(k)}\|_{\Omega}^2.
\end{eqnarray*}
For HoRMP-LS, we have
\begin{eqnarray*}
\|\mathcal R^{(k+1)}\|_F^2 &=& \min_{(\alpha_1,\alpha_2)}\|   \alpha_1\mathcal W^{(k)} + \alpha_2 \mathcal S^{(k)}-\mathcal B\|_{\Omega}^2\\
&\leq& \min_{\alpha}\|  \mathcal W^{(k)} -\mathcal B + \alpha\mathcal S^{(k)}\|_{\Omega}^2 = \min_{\alpha}\|\mathcal R^{(k)} + \alpha\mathcal S^{(k)}\|_{\Omega}^2. 
\end{eqnarray*}
It follows that for HoMP-LS, it naturally holds $\|\mathcal R^{(k+1)}\|_F^2  = \min_{\alpha}\|\mathcal R^{(k)} + \alpha\mathcal S^{(k)}\|_{\Omega}^2$. Therefore, we can uniformly analyze the upper bound in terms of $\min_{\alpha}\|\mathcal R^{(k)} + \alpha\mathcal S^{(k)}\|_{\Omega}^2$. We have
\begin{eqnarray*}
\min_{\alpha}\|\mathcal R^{(k)} + \alpha\mathcal S^{(k)}\|_{\Omega}^2 = \|\mathcal R^{(k)}\|_F^2 - \frac{\langle \mathcal R^{(k)},\mathcal \mathcal S^{(k) }\rangle_{\Omega}^2 }{\|\mathcal S^{(k)}\|_{\Omega}^2}\leq   \|\mathcal R^{(k)}\|_F^2 - \rho^2 \|\mathbf  R^{(k)}_{(1,2)}\|_2^2 \leq \left(1- \frac{\rho^2}{ {n_1n_2}}\right)\|\mathcal R^{(k)}\|_F^2,
\end{eqnarray*} 
where the first inequality follows from $\langle   \mathcal R^{(k)},\mathcal S^{(k)}\rangle_{\Omega} =   \langle \nabla F(\mathcal W^{(k)}),\mathcal S^{(k)}\rangle\geq \rho \|\nabla F(\mathcal W^{(k)})_{(1,2)}\|_2^2 = \rho\|\mathbf R^{(k)}_{(1,2)}\|_2^2$, and the last inequality is due to the relationship between the spectral norm and the Frobenius norm.  
\end{IEEEproof} 
 
\subsubsection{Multilinear multitask learning}\label{sec:mml_ls}
In the least squares sense, the cost function of multilinear multitask learning is given by
$$F(\mathcal W) =  {1}/{2}\sum^T_{t=1} {m_t^{-1}}\sum^{m_t}_{i=1} (\langle \mathbf x^t_i,\mathbf w^t\rangle-y^t_i)^2.$$
Letting $\mathbf X^t\in\mathbb R^{m_t\times D}$ be formed by stacking the samples (transposed to rows) corresponding to the $t$-th task row by row, i.e., $(\mathbf X^t)^{\top} =    [\mathbf x^t_1,\ldots,\mathbf x^t_{m_t}]$, we derive a compact form of   $F $ 
$$F(\mathcal W) =  1/2\sum^T_{t=1} { m_t^{-1}}\|\mathbf X^t\mathbf w^t - \mathbf y^t\|_F^2,$$
and the corresponding gradient, which is represented by the mode-$1$ unfolding of $\nabla F(\mathcal W)$,   can be written as
\begin{equation}\label{eq:lc_ls_1}
\nabla F(\mathcal W)_{(1)} =  \left[(\mathbf X^1)^{\top}(\mathbf X^1\mathbf w^1-\mathbf y^1)/m_1,\ldots, (\mathbf X^T)^{\top}(\mathbf X^T\mathbf w^T-\mathbf y^T)/m_T  \right].
\end{equation}
We have the following results.
\begin{theorem}[Linear convergence rate for MLMTL] \label{th:lc_mml_ls}
 Assume that the matrices $  \mathbf X^t(\mathbf X^t)^{\top}$ are all positive definite, $1\leq t\leq T$, whose smallest eigenvalues are uniformly lower bounded by $\lambda_{\min}$, while the largest eigenvalues are uniformly upper bounded by $\lambda_{\max}$. Let   $\mathcal S^{(k)}$ be generated by Subroutine \ref{sub:2} with input $\nabla F(\mathcal W^{(k)})$, and \eqref{eq:key_inequality} holds. Let $m_{\max}:=\max_{1\leq t\leq T}m_t$. If $\{\mathcal W^{(k)} \}$ is generated by HoMP-LS, HoRMP-LS or HoOMP-LS,  then there holds
$$F(\mathcal W^{(k+1)})\leq   \left( 1-  \frac{\rho \lambda_{\min}m_{\max}^{-1} }{ n_1n_2 \lambda_{\max} \sum^T_{t=1}m_t^{-1} } \right) F(\mathcal W^{(k)}).$$
\end{theorem}
\begin{IEEEproof}
We denote $\mathbf r^{t,(k)}:= \mathbf X^t\mathbf w^{t,(k)} - \mathbf y^t$, $1\leq t\leq T$. We first consider HoOMP-LS. Similar to the previous analysis, we have
\begin{eqnarray*}
F(\mathcal W^{(k+1)}) &=&  \sum^T_{t=1}\frac{1}{2m_t}\|\mathbf r^{t,(k+1)}\|_F^2 =  \sum^T_{t=1}\frac{1}{2m_t}\|\mathbf X^t\mathbf w^{t,(k+1)} - \mathbf y^{t}\|_F^2\\ &=& \min_{\boldsymbol{\alpha}}  \sum^T_{t=1}\frac{1}{2m_t}\|\mathbf X^t(\sum^k_{i=0}\alpha_i\mathbf s^{t,(i)}) - \mathbf y^{t}\|_F^2\\
&\leq& \min_{ {\alpha}} \sum^T_{t=1}\frac{1}{2m_t}\|\mathbf X^t(\sum^{k-1}_{i=0}\overline{\alpha^k_i}\mathbf s^{t,(i)} + \alpha \mathbf s^{t,(k)}) - \mathbf y^{t}\|_F^2\\
&=&  \min_{ {\alpha}} \sum^T_{t=1}\frac{1}{2m_t}\|\mathbf X^t\mathbf w^{t,(k)}-\mathbf y^t + \alpha\mathbf X^t\mathbf s^{t,(k)}\|_F^2 =  \min_{ {\alpha}}\sum^T_{t=1}\frac{1}{2m_t}\|\mathbf r^{t,(k)} + \alpha\mathbf X^t\mathbf s^{t,(k)}\|_F^2.
\end{eqnarray*}
We then consider HoRMP-LS as follows
\begin{eqnarray*}
F(\mathcal W^{(k+1)}) &=&  \sum^T_{t=1}\frac{1}{2m_t}\|\mathbf r^{t,(k+1)}\|_F^2 =  \sum^T_{t=1}\frac{1}{2m_t}\|\mathbf X^t\mathbf w^{t,(k+1)} - \mathbf y^{t}\|_F^2\\
&=& \min_{(\alpha_1,\alpha_2)} \sum^T_{t=1}\frac{1}{2m_t}\|\mathbf X^t(\alpha_1\mathbf w^{t,(k)}+\alpha_2\mathbf s^{t,(k)}) - \mathbf y^{t}\|_F^2\\
&\leq&  \min_{ {\alpha}} \sum^T_{t=1}\frac{1}{2m_t}\|\mathbf X^t\mathbf w^{t,(k)}-\mathbf y^t + \alpha\mathbf X^t\mathbf s^{t,(k)}\|_F^2 =  \min_{ {\alpha}} \sum^T_{t=1}\frac{1}{2m_t}\|\mathbf r^{t,(k)} + \alpha\mathbf X^t\mathbf s^{t,(k)}\|_F^2.
\end{eqnarray*}
And it naturally holds $F(\mathcal W^{(k+1)}) = \min_{ {\alpha}} \sum^T_{t=1}\frac{1}{2m_t}\|\mathbf r^{t,(k)} + \alpha\mathbf X^t\mathbf s^{t,(k)}\|_F^2$ for HoMP-LS.   We then analyze the upper bound of  $\min_{ {\alpha}} \sum^T_{t=1}\frac{1}{2m_t}\|\mathbf r^{t,(k)} + \alpha\mathbf X^t\mathbf s^{t,(k)}\|_F^2$. From the optimality condition
$$
\sum^T_{t=1}\frac{1}{m_t}(\mathbf X^t\mathbf s^{t,(k)})^{\top}(\mathbf r^{t,(k)} +\alpha \mathbf X^t\mathbf s^{t,(k)})=0
$$
 we get
\begin{eqnarray*}
\alpha &=& -\frac{\sum^T_{t=1}\frac{1}{m_t}\langle  \mathbf X^t\mathbf s^{t,(k)}, \mathbf r^{t,(k)}\rangle}{\sum^T_{t=1}m_t^{-1}\|\mathbf X^t\mathbf s^{t,(k)}\|_F^2} 
\end{eqnarray*}
and so
\begin{equation}\label{eq:lc_ls:1}
F(\mathcal W^{(k+1)})\leq \min_{ {\alpha}} \sum^T_{t=1}\frac{1}{2m_t}\|\mathbf r^{t,(k)} + \alpha\mathbf X^t\mathbf s^{t,(k)}\|_F^2  =  \sum^T_{t=1}\frac{1}{2m_t}\|\mathbf r^{t,(k)}\|_F^2 -  \frac{(\sum^T_{t=1}{m_t^{-1}}\langle \mathbf X^t\mathbf s^{t,(k)}, \mathbf r^{t,(k)} \rangle)^2 }{2\sum^T_{t=1}m_t^{-1}\|\mathbf X^t\mathbf s^{t,(k)}\|_F^2}.
\end{equation}
Recalling the definition of $\mathbf r^{t,(k)}$ and $\nabla F(\mathcal W^{(k)})$, the numerator of the second term is exactly $\langle \nabla F(\mathcal W^{(k)}),\mathcal S^{(k)}\rangle^2$. Using the inequalities 
$$\langle \nabla F(\mathcal W^{(k)}),\mathcal S^{(k)}\rangle \geq \rho \|\nabla F(\mathcal W^{(k)})_{(1,2)}\|_2^2 \geq \frac{\rho}{n_1n_2}\|\nabla F(\mathcal W^{(k)})\|_F^2,$$
and from the assumption that  $\|\mathbf X^t\mathbf s^{t,(k)}\|_F^2\leq \lambda_{\max}\|\mathbf s^{t,(k)}\|_F^2$ and $\|(\mathbf X^t)^{\top}\mathbf r^{t,(k)}\|_F^2\geq \lambda_{\min}\|\mathbf r^{t,(k)}\|_F^2$,  the second term of \eqref{eq:lc_ls:1} can be lower bounded as follows
\begin{eqnarray*}
\frac{(\sum^T_{t=1}{m_t^{-1}}\langle \mathbf X^t\mathbf s^{t,(k)}, \mathbf r^{t,(k)} \rangle)^2}{2\sum^T_{t=1}m_t^{-1}\|\mathbf X^t\mathbf s^{t,(k)}\|_F^2} & \geq &  \frac{\rho}{2n_1n_2}\frac{ \|\nabla F(\mathcal W^{(k)}) \|_F^2 }{  \sum^T_{t=1}m_t^{-1}\|\mathbf X^t\mathbf s^{t,(k)}\|_F^2 }\\
&\geq&  \frac{\rho \lambda_{\min}}{2n_1n_2 \lambda_{\max}} \frac{  \sum^T_{t=1}m_t^{-2}\|  \mathbf r^{t,(k)}\|_F^2   }{  \sum^T_{t=1}m_t^{-1}  \|\mathbf s^{t,(k)}\|_F^2  }\\
&\geq& \frac{\rho \lambda_{\min}m_{\max}^{-1} }{ n_1n_2 \lambda_{\max} \sum^T_{t=1}m_t^{-1} }    \sum^T_{t=1}\frac{1}{2m_t}\|  \mathbf r^{t,(k)}\|_F^2.   
\end{eqnarray*}
This together with \eqref{eq:lc_ls:1} yields
$$F(\mathcal W^{(k+1)})\leq \left( 1-  \frac{\rho \lambda_{\min}m_{\max}^{-1} }{ n_1n_2 \lambda_{\max} \sum^T_{t=1}m_t^{-1} } \right) \sum^T_{t=1}\frac{1}{2m_t}\|  \mathbf r^{t,(k)}\|_F^2 = \left( 1-  \frac{\rho \lambda_{\min}m_{\max}^{-1} }{ n_1n_2 \lambda_{\max} \sum^T_{t=1}m_t^{-1} } \right) F(\mathcal W^{(k)}),$$
as desired.
\end{IEEEproof}

In real world applications, however, the assumption that every matrix $\mathbf X^t(\mathbf X^t)^{\top}$ is positive definite may not hold, e.g., when the sample size $m_t$ is larger than the size of the feature space $D$. To establish the linear convergence in this setting, we consider the following two cases: 

1)  $(\mathbf X^t)^{\top}\mathbf X^t$ is positive semidefinite; we add the $L_2$ regularization to the cost function, i.e., now the new cost function is
$$\hat F(\mathcal W) = F(\mathcal W) + \lambda \sum^T_{t=1}\frac{1}{2m_t}\|\mathbf w^t\|_F^2,$$
where $\lambda>0$ is a regularization parameter.
In this case, we rewrite $\hat F$ as
\begin{eqnarray*}
\hat F(\mathcal W) &=& \hat G(\mathcal W) + C\\
&=&\sum^T_{t=1}\frac{1}{2m_t} \|\mathbf{\hat X}^t\mathbf w^t - \mathbf z^t\|_F^2 + C,
\end{eqnarray*}
where $\mathbf{\hat X}^t \in\mathbb R^{D\times D}$ is the square root of $(\mathbf X^t)^{\top}\mathbf X^t + \lambda I$ and is a symmetric matrix,   $\mathbf z^t = (\mathbf{\hat X}^t)^{-1}(\mathbf X^t)^{\top}\mathbf y^t$, and $C:=\sum^T_{t=1}(2m_t)^{-1}(\|\mathbf y^t\|_F^2 - \|\mathbf z^t\|_F^2 )$ denotes the constant term. Now $ \mathbf{\hat X}^t(\mathbf{\hat X}^t)^{\top}$ is positive definite, which meets the assumption of Theorem \ref{th:lc_mml_ls}. 

2) $(\mathbf X^t)^{\top}\mathbf X^t$ is positive  definite; we simply set $\lambda = 0$ above, and also obtain that  $ \mathbf{\hat X}^t(\mathbf{\hat X}^t)^{\top}$ is positive definite.

Therefore, we have linear convergence rate in the sense of $\hat G(\cdot)$.
\begin{corollary}[Linear convergence for MLMTL without the positive definiteness assumption] \label{th:lc_mml_1_ls}
Assume that $\mathcal W\in\mathbb T$. Let $\hat F(\cdot)$, $\hat G(\cdot)$ and $\mathbf{\hat X}^t$ be defined as above. Assume that the smallest and the largest eigenvalues of matrices $(\mathbf {\hat X}^t)^2  $ are respectively   lower bounded by $\lambda_{\min}$ and upper bounded by $\lambda_{\max}$. Let $0<\rho\leq 1$ be defined as in \eqref{eq:key_inequality}, and let $m_{\max}:=\max_{1\leq t\leq T}m_t$. If $\{\mathcal W^{(k)} \}$ is generated by HoMP-LS, HoRMP-LS or HoOMP-LS with the cost function given by $\hat F(\cdot)$,  then there holds
$$\hat G(\mathcal W^{(k+1)})\leq   \left( 1-  \frac{\rho \lambda_{\min}m_{\max}^{-1} }{ n_1n_2 \lambda_{\max} \sum^T_{t=1}m_t^{-1} } \right) \hat G(\mathcal W^{(k)}).$$
\end{corollary}

 \subsection{A class of loss functions}\label{sec:lc_non_ls}
 
\begin{figure}[H]
   \centering
      \begin{subfigure}[b]{0.5\textwidth}
                 \includegraphics[width=\textwidth]{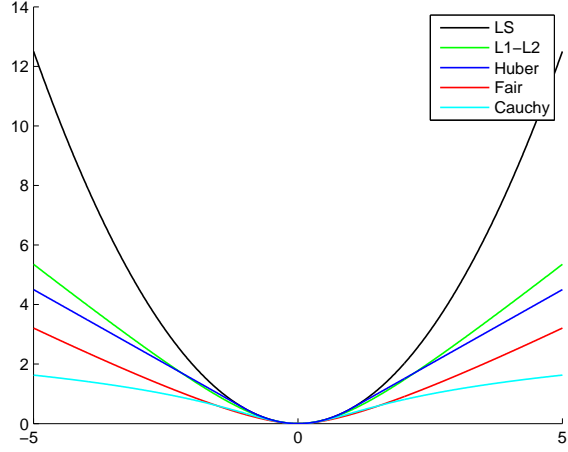}\caption{Different loss functions.}
                              \end{subfigure}  
       \begin{subfigure}[b]{0.5\textwidth}
                  \includegraphics[width=\textwidth]{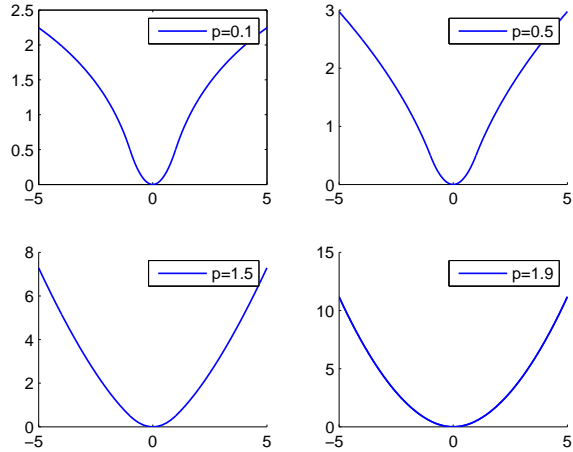}\caption{Generalized Huber's loss functions with different $p$ values, with $\delta=1$.}
                               \end{subfigure}                              
                                      \caption{Different loss functions. }  \label{fig:loss}               
\end{figure}

 In this subsection, we conduct the convergence rate analysis for a class  of loss functions which may be   nonconvex. To this end, we first present some assumptions that characterize such class of   functions $\ell(t)$:
 \begin{assumption}\label{ass:2}
 \begin{enumerate}
 \item $\ell(t)\geq 0$, $\ell(0)=0$, $\ell(t)=\ell(-t)$, $\ell(t)\leq  t^2/2,~\forall t\in\mathbb R$, and $\ell(t)$ is coercive;
 \item $\ell(t)$ is continuously differentiable;
 \item $ |\ell^{'}(s) - \ell^{'}(t)|\leq |s-t|$, $\forall s,t \in\mathbb R$;
 \item denote   $\psi(t):= \ell^{'}(t)/t$; then $0\leq \psi(t)\leq 1$, $\psi(-t)=\psi(t)$, and $\psi(0)$ exists and is finite;
 \item $\psi(t)\not\rightarrow 0$ if $t\not\rightarrow \infty$.
 \end{enumerate}
 \end{assumption}
 The above assumptions are not restrictive, as they are met for a variety of loss functions, e.g., the Huber's loss \cite{huber2011robust}, the $L_1-L_2$ loss $2(\sqrt{1 + t^2/2}-1)$, the Fair loss $  \sigma^2(|t|/\sigma - \log(1+ |t|/\sigma) )$, and the Cauchy loss $ \sigma^2/2\log(1+ t^2/\sigma^2)$. Here $\sigma$ is a parameter.  Note that the Cauchy loss   is nonconvex. 
 We can also define the generalized Huber's   functions that satisfies Assumption \ref{ass:2}:  
$$
 \ell(t):= \left\{ \begin{array}{ll}
  t^2/2& |t|\leq \delta \\
  \delta^{2-p}(|t|^p/p + \delta^p/2 - \delta^p/p)& |t|>\delta,
 \end{array} \right.
$$
where $\delta>0$ is a parameter and $0<p\leq 2$. When $p=1$ it reduces to the Huber's loss; when $p=2$ it is exactly the least squares loss. If $p<1$ then this class of functions is also nonconvex. Fig. \ref{fig:loss} plots the losses mentioned above.


We still begin our analysis from the tensor completion setting.
\subsubsection{Tensor completion with a general loss}
 With the loss function $\ell(t)$ satisfying Assumption \ref{ass:2}, the cost function of tensor completion is given by
\begin{equation}\label{cost:tc_non_ls}
F(\mathcal W) = \sum_{(i_1,\ldots,i_N)\in \Omega}\ell( \mathcal W_{i_1\cdots i_N} -\mathcal B_{i_1\cdots i_N}). 
\end{equation}
 That is to say, we penalize the tensors entry by entry with $\ell(\cdot)$. Accordingly, the gradient of $F(\cdot )$ at $\mathcal W$ can be derived as
 \begin{equation}\label{eq:non_ls:2}
\nabla F(\mathcal W) =    \Psi_{\Omega}\circ(\mathcal W-\mathcal B)_{\Omega} ,
 \end{equation}
 where $\circ$ denotes the Hadamard operator, i.e., entry-wise product, and $\Psi$ is defined as $\Psi_{i_1\cdots i_N} = \psi(\mathcal R_{i_1\cdots i_N} )$ where we denote $\mathcal R:=\mathcal W-\mathcal B$, and $\psi$ is defined in Assumption \ref{ass:2}. In robust statistics \cite{huber2011robust}, with an appropriate loss such as those mentioned below Assumption \ref{ass:2}, each entry of $\Psi$ can be seen as a weight assigned to the corresponding entry of $\mathcal R $.   As the entry of $\mathcal R$ goes large, the corresponding entry of $\Psi$ will decrease to control the influence of the large deviations. 
 
If the cost function of tensor completion is given by \eqref{cost:tc_non_ls}, then one can use Assumption \ref{ass:2}--3) to verify that
\begin{equation}\label{eq:non_ls:1}
\|\nabla F(\mathcal U) -\nabla F(\mathcal V) \|_F \leq \|\mathcal U-\mathcal V\|_F,
\end{equation}
i.e., the gradient of $F(\cdot)$ is Lipschitz continuous with constant being $1$. Therefore, 
we can apply strategy 4) of Algorithm \ref{alg:mp}, HoMP-G to solve it. In the following, we present our results and analysis for tensor completion solved by HoMP-G. 
 \begin{theorem}[Linear convergence for tensor completion with   a general loss] \label{th:lc_tc_non_ls}
  Assume that the cost function is given by \eqref{cost:tc_non_ls} with a loss function $\ell(\cdot)$ satisfying Assumption \ref{ass:2}. 
  Let $\rho$ be defined as in \eqref{eq:key_inequality}. If $\{\mathcal W^{(k)} \}$ is generated by HoMP-G, then there holds
  $$F(\mathcal W^{(k+1)}) \leq \left (1-  \frac{\rho q^2}{n_1n_2} \right)F(\mathcal W^{(k)}),$$
  where $0<q\leq 1$ is  some constant. 
 \end{theorem}
  
  \begin{IEEEproof}
  Let $\mathcal W^{(k+1)} = \mathcal W^{(k)} + \overline{\alpha}\mathcal S^{(k)}$ where $\overline \alpha = -\langle \nabla F(\mathcal W^{(k)}),\mathcal S^{(k)}\rangle $.  According to \eqref{eq:non_ls:1}, we have
  \begin{eqnarray*}
  F(\mathcal W^{(k+1)})&\leq& F(\mathcal W^{(k)}) + \langle \nabla F(\mathcal W^{(k)}),\mathcal W^{(k+1)}-\mathcal W^{(k)}\rangle + 2^{-1}\|\mathcal W^{(k+1)} - \mathcal W^{(k)}\|_F^2\\
  &=& F(\mathcal W^{(k )}) - 2^{-1} {\langle \nabla F(\mathcal W^{(k)}),\mathcal S^{(k)}\rangle^2} \\
  &\leq& F(\mathcal W^{(k)}) - \frac{{\rho}}{2n_1n_2}\|\nabla F(\mathcal W^{(k )})\|_F^2.
  \end{eqnarray*}
Therefore, $\{F(\mathcal W^{(k)}) \}$ is nonincreasing. From the definition of $F(\cdot)$ and Assumption \ref{ass:2}--1),  it follows that $\{\mathcal W^{(k)}_{\Omega} \}$ is uniformly bounded.  We consider the term $\|\nabla F(\mathcal W^{(k )})\|_F^2$. From \eqref{eq:non_ls:2}, it follows
  \begin{eqnarray*}
\|\nabla F(\mathcal W^{(k )})\|_F^2 = \sum_{(i_1,\ldots,i_N)\in\Omega} \psi(\mathcal R^{(k)}_{i_1,\ldots,i_N})^2 (\mathcal R^{(k)}_{i_1,\ldots,i_N})^2,
  \end{eqnarray*}
where $\mathcal R^{(k)} = \mathcal W^{(k)}_{\Omega} - \mathcal B_{\Omega}$. From Assumption \ref{ass:2}--5), the boundedness of $\{\mathcal W^{(k)}_{\Omega} \}$ and $\mathcal B_{\Omega}$ implies that $\psi(\mathcal R^{(k)}_{i_1,\ldots,i_N})$ is   uniformly lower bounded away from zero for all $k\geq 0$ and for all $(i_1,\ldots,i_N)\in\Omega$. Without loss of generality we assume that $\psi(\mathcal R^{(k)}_{i_1,\ldots,i_N})\geq q>0$, where the magnitude of $q$ only depends on the magnitude of $\mathcal B_{\Omega}$. On the other hand, Assumption \ref{ass:2}--1) tells us that $  (\mathcal R^{(k)}_{i_1,\ldots,i_N})^2 \geq 2\ell(\mathcal R^{(k)}_{i_1,\ldots,i_N})$. As a consequence,
$$\|\nabla F(\mathcal W^{(k+1)})\|_F^2  \geq 2q^2 F(\mathcal W^{(k)}) ,$$ 
and so
$$F(\mathcal W^{(k+1)}) \leq \left (1-  \frac{\rho q^2}{n_1n_2} \right)F(\mathcal W^{(k)}),$$
as desired.
  \end{IEEEproof}
  
  \subsubsection{Mutilinear multitask learning with a general loss}
 In this setting, the cost function is given by
 \begin{equation}\label{cost:mml_non_ls}
F(\mathcal W) = \sum^T_{t=1} {m_t^{-1}}\sum^{m_t}_{i=1}\ell( \langle \mathbf x^t_i,\mathbf w^t\rangle - y^t_i).
 \end{equation}
 That is, the noise or outliers are penalized sample-wisely by using $\ell(\cdot)$. The gradient of $F(\cdot)$ at $\mathcal W$ can be derived as follows: its gradient at $\mathbf w^t$ is given by $$(\mathbf X^t)^{\top}\boldsymbol{\Lambda}^t(\mathbf X^t\mathbf w^t-\mathbf y^t)/m_t,$$ where $\mathbf X^t$ is the same as that defined in subsection \ref{sec:mml_ls}, and $\boldsymbol{\Lambda}^t\in\mathbb R^{m_t\times m_t}$ is a diagonal matrix, whose $i$-th diagonal entry is $\boldsymbol{\Lambda}^t_{ii} = \psi(\langle \mathbf x^t_i,\mathbf w^t\rangle - y^t_i)$. Therefore, the gradient of $F(\cdot)$ at $\mathcal W$ in terms of its mode-$1$ unfolding   can be written as
 \begin{equation}\label{eq:lc_non_ls_1}
 \nabla F(\mathcal W)_{(1)} =  \left[(\mathbf X^1)^{\top}\boldsymbol{\Lambda}^1(\mathbf X^1\mathbf w^1-\mathbf y^1)/m_1,\ldots, (\mathbf X^T)^{\top}\boldsymbol{\Lambda}^T(\mathbf X^T\mathbf w^T-\mathbf y^T)/m_T  \right].
 \end{equation}
 The difference  between \eqref{eq:lc_non_ls_1} and its least squares counterpart
\eqref{eq:lc_ls_1} is the matrix $\boldsymbol{\Lambda}^t$, which acts as a weight matrix to remove the large deviation between $\mathbf X^t\mathbf w^t$ and $\mathbf y^t$ if necessary. Using Assumption \ref{ass:2}--3), one can verify that its gradient is Lipschitz continuous,
\begin{equation}\label{eq:mml_non_ls:1}
\|\nabla F(\mathcal U) - \nabla F(\mathcal V)\|_F  \leq   \left(\sum^T_{t=1}\|\mathbf X^t\|_2^4 \|\mathbf u^t-\mathbf v^t\|_F^2\right )^{1/2} \leq \lambda_{\max}\|\mathcal U-\mathcal V\|_F, 
\end{equation} 
 where we use $\lambda_{\max} = \max_{1\leq t\leq T}\|\mathbf X^t\|_2^2$ as a Lipschitz constant. Thus HoMP-G can also be applied.
 We present our convergence rate results in the following.
 \begin{theorem}[Linear convergence for multilinear multitask learning with   a general loss] \label{th:lc_mml_non_ls}
   Assume that the cost function is given by \eqref{cost:mml_non_ls} with a loss function $\ell(\cdot)$ satisfying Assumption \ref{ass:2}. Assume that the matrices $\mathbf X^t(\mathbf X^t)^{\top}$ are all positive definite, $1\leq t\leq T$, whose smallest eigenvalues are uniformly lower bounded by $\lambda_{\min}$, while the largest eigenvalues are uniformly upper bounded by $\lambda_{\max}$.  
 Let $\rho$ be defined as in \eqref{eq:key_inequality},  and let $m_{\max}:=\max_{1\leq t\leq T}m_t$.  If $\{\mathcal W^{(k)} \}$ is generated by HoMP-G, then there holds
    $$F(\mathcal W^{(k+1)}) \leq \left(1- \frac{\rho\lambda_{\min}  m_{\max}^{-1}q^2 }{ n_1n_2 \lambda_{\max}} \right)F(\mathcal W^{(k)}),$$
   where $0<q\leq 1$ is a some constant.
  \end{theorem}
  
  \begin{IEEEproof} We denote $\mathbf r^{t,(k)}:= \mathbf X^t\mathbf w^{t,(k)} - \mathbf y^t$, $1\leq t\leq T$.
According to the strategy MP-non-LS in Algorithm \ref{alg:mp}, we let $\mathcal W^{(k+1)} = \mathcal W^{(k)} + \overline{\alpha}\mathcal S^{(k)}$, where $$\overline \alpha = -\langle \nabla F(\mathcal W^{(k)}),\mathcal S^{(k)}\rangle/ \lambda_{\max}  .$$   Using \eqref{eq:mml_non_ls:1}  and similar to the proof of Theorem \ref{th:lc_tc_non_ls}, we have
  \begin{eqnarray*}
  F(\mathcal W^{(k+1)})&\leq& F(\mathcal W^{(k)}) + \langle \nabla F(\mathcal W^{(k)}),\mathcal W^{(k+1)}-\mathcal W^{(k)}\rangle + \frac{\lambda_{\max}}{2}\|\mathcal W^{(k+1)} - \mathcal W^{(k)}\|_F^2\\
  &=& F(\mathcal W^{(k )}) - \frac{1}{2\lambda_{\max}} {\langle \nabla F(\mathcal W^{(k)}),\mathcal S^{(k)}\rangle^2} \\
  &\leq& F(\mathcal W^{(k)}) - \frac{{\rho}}{2n_1n_2\lambda_{\max}}\|\nabla F(\mathcal W^{(k )})\|_F^2.
  \end{eqnarray*}  
 It follows that $\{F(\mathcal W^{(k)}\}$ is a nonincreasing sequence. This together with the definition of $F(\cdot)$ and Assumption \ref{ass:2}--1) implies that all the sequence $\{\mathbf r^{t,(k)} \}$ are uniformly bounded, $1\leq t\leq T$. Recalling Assumption \ref{ass:2}--5) and recalling $\boldsymbol{\Lambda}^t_{ii} = \psi(\langle \mathbf x^t_i,\mathbf w^t\rangle - y^t_i)$, the boundedness of $\{\mathbf r^{t,(k)} \}$ also implies that there is a universal constant $q>0$ such that $\boldsymbol{\Lambda}^t_{ii} 
 \geq q$ for $1\leq i\leq m_t$, $1\leq t\leq T$, and $k
\geq 0$, where the magnitude of $q$ only depends on the magnitude of the samples $\{\mathbf x^t_i,y^t_i \}$. Now we can
 consider the term $\|\nabla F(\mathcal W^{(k )})\|_F^2$. From \eqref{eq:lc_non_ls_1}, it follows
    \begin{eqnarray*}
  \|\nabla F(\mathcal W^{(k )})\|_F^2 &=& \sum^T_{t=1} m_t^{-2} \| (\mathbf X^t)^{\top}\boldsymbol{\Lambda}^t(\mathbf X^t\mathbf w^{t,(k)}-\mathbf y^t)\|_F^2\\
  &\geq&  \lambda_{\min}m_{\max}^{-1}\sum^T_{t=1}  m_{t}^{-1} \|\boldsymbol{\Lambda}^t(\mathbf X^t\mathbf w^{t,(k)} - \mathbf y^t)\|_F^2\\
  &\geq& \lambda_{\min}m_{\max}^{-1} q^2\sum^T_{t=1} m_{t}^{-1} \|\mathbf r^{t,(k)}\|_F^2\\
  &\geq& \lambda_{\min}m_{\max}^{-1} q^2\sum^T_{t=1} m_{t}^{-1}\sum^{m_t}_{i=1}\ell( \langle \mathbf x^t_i,\mathbf w^t\rangle - y^t_i )\\
  &=&   2\lambda_{\min}m_{\max}^{-1} q^2 F(\mathcal W^{(k)}),
    \end{eqnarray*}
    where the first inequality is due to $\|\mathbf X^t\|_2^2\geq\lambda_{\min}$, and the last inequality follows from Assumption \ref{ass:2}--1), and so
    $$F(\mathcal W^{(k+1)}) \leq \left(1- \frac{\rho\lambda_{\min}  m_{\max}^{-1}q^2 }{ n_1n_2 \lambda_{\max}} \right)F(\mathcal W^{(k)}).$$
    The proof is completed.
  \end{IEEEproof}
  
  To study the case that the matrices $\mathbf X^t(\mathbf X^t)^{\top}$ may not be all positive definite, we also make some modifications to the cost function as that discussed in subsection \ref{sec:mml_ls}. We let $(\mathbf{\hat X}^t)^2 := (\mathbf X^t)^{\top}\mathbf X^t + \lambda I$, where $\lambda>0$ if $(\mathbf X^t)^{\top}\mathbf X^t$ is positive semidefinite, and $\lambda = 0$ if $(\mathbf X^t)^{\top}\mathbf X^t$ is positive definite. We   denote $\mathbf z^t = (\mathbf{\hat X}^t)^{-1}(\mathbf X^t)^{\top}\mathbf y^t$, and reconstruct the cost function as 
  \begin{equation}\label{eq:cost3}
 \hat G(\mathcal W) =  \sum^T_{t=1}m_t^{-1}\sum^{m_t}_{i=1}\ell( \langle \mathbf{\hat x}^t_{i}, \mathbf w^t\rangle -z^t_i),
  \end{equation}
  where $\mathbf {\hat x^t}_i\in\mathbb R^{D}$ is the $i$-th column of $\mathbf {\hat X}^t$. Similar to Corollary \ref{th:lc_mml_1_ls}, we have
  
  \begin{corollary}[Linear convergence for multilinear
  multitask learning with a general loss and without the positive definiteness assumption] \label{th:lc_mml_1_non_ls}  Assume that the cost function is given by \eqref{eq:cost3} with a loss function $\ell(\cdot)$ satisfying Assumption \ref{ass:2}.
   Let   $\hat G(\cdot)$ and $\mathbf{\hat X^t}$ be defined as above. Assume that the smallest and the largest eigenvalues of matrices $(\mathbf {\hat X^t})^2$ are respectively   lower bounded by $\lambda_{\min}$ and upper bounded by $\lambda_{\max}$.  Let $\rho$ be defined as in \eqref{eq:key_inequality}, and let $m_{\max}:=\max_{1\leq t\leq T}m_t$. If $\{\mathcal W^{(k)} \}$ is generated by   HoMP-G,  then there holds
  $$\hat G(\mathcal W^{(k+1)})\leq    \left(1- \frac{\rho\lambda_{\min} m_{\max}^{-1}q^2 }{n_1n_2\lambda_{\max} } \right) \hat G(\mathcal W^{(k)}).$$
  \end{corollary}
  
\begin{remark}
Before ending this section, we remark that 1) inequality \eqref{eq:mp:sigular} is very important in deriving the linear convergence. Without it, only sublinear convergence rate can be obtained; 2) all the convergence rates obtained in this section can be improved in theory by using, e.g., the methods in \cite{So2010,hjlz2012} to improve the ratio $\rho$ in \eqref{eq:key_inequality}, while getting less efficiency.
\end{remark}  
  
\section{Numerical Experiments}\label{sec:exp}
 In this section, we present some  numerical experiments on synthetic data as well as real data, focusing on the applications: (robust) tensor completion and MLMTL.  All the numerical computations are conducted on an Intel i7-3770 CPU desktop computer with 16 GB of RAM. The supporting software is MATLAB R2013a.  
 
 \subsection{Tensor completion}
 
 Our HoMP-type methods with least squares loss are compared with $4$ state-of-the-art methods: generalized conditional gradient (GCG) \cite{yu2014approximate}, HaLRTC \cite{liu2013tensor}, factor priors (FP) \cite{chen2014simultaneous} and TMac \cite{xu2013parallel}. GCG also solves an approximate tensor singular value problem \eqref{eq:mp:sigular} at each iteration; HaLRTC solves a convex relaxation of tensor completion with sum of matrix nuclear norms; FP uses some prior knowledges of the tensors, and TMac is based on tensor factorization. 
 The stopping criterion for HoMPS is when the residual $\mathcal R^{(k)}$  used in Sect. \ref{sec:lc} is less than a threshold. For all the methods except HaLRTC, the threshold of the stopping criterion is $\epsilon = 10^{-5}$; for HaLRTC, we  set  $10^{-6}$ because otherwise it cannot generate a good result. The max iteration for all the methods is $500$, whereas for HoMPs, the max iteration $K$ is the only parameter, which is tuned by 10-fold cross validation over $K\in\{100,200,300,400,500\}$. All the results are averaged over ten instances.
 
 \subsubsection{Synthetic data}
\begin{figure}[H]
   \centering
      \begin{subfigure}[b]{0.5\textwidth}
                 \includegraphics[width=\textwidth]{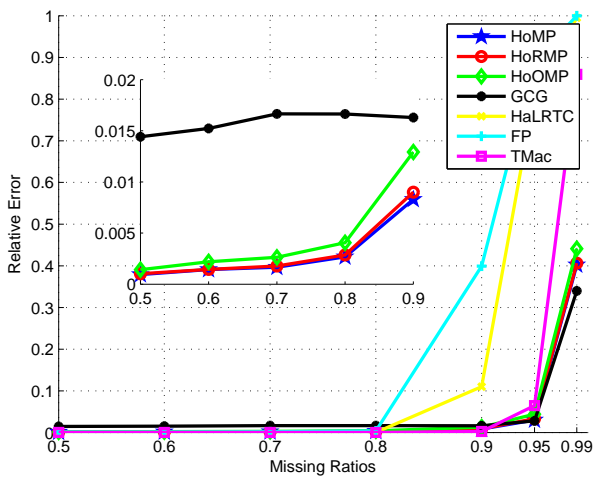}
                              \end{subfigure}          \caption{Tensor completion  results on synthetic data ($200\times200\times200$, CP-rank $=10$) in terms of relative error. }  \label{fig:results_tc_syn}               
\end{figure}

Third order tensors of size  $200\times200\times200$ are randomly generated, with CP-rank $10$. Some entries are randomly missing, with missing ratio (MR) varies in $\{0.5,0.6,0.7,0.8,0.9,0.95,0.99\}$. The relative error $\|\mathcal X^*-\mathcal B\|_F/\|\mathcal B\|_F$ and the computational time (in Second) are respectively reported in Fig. \ref{fig:results_tc_syn} and Table \ref{tab:results_tc_syn}. The results in Fig. \ref{fig:results_tc_syn} show that HoMPs and GCG have better performances, particularly when the MR value is very high. The results   in the subfigure of Fig. \ref{fig:results_tc_syn} show that HoMPs perform better than GCG when the MR values in $[0.5,0.9]$. Table \ref{tab:results_tc_syn} shows that HoMPs are more efficient than other methods, particularly when the MR value is very high. That is because we have optimized our   codes by utilizing the sparsity of the data and using the sparsity manipulation in Matlab. Comparing between the three HoMPs, it is interesting to see that the relative error of HoOMP is not as good as the other two. That may be because HoOMP learns the tensor more greedily and leads to overfitting. And HoOMP is the slowest among the three  methods, as it has to solve a larger linear equation system to obtain the weights.

\begin{table}[H]  
\renewcommand{\arraystretch}{0.5}
  \setlength{\tabcolsep}{3.5pt}
  \centering
  \caption{Efficiency comparison of different methods on tensor completion on synthetic data ($200\times200\times200$, CP-rank $=10$).}
    \begin{tabular}{rrrrrrrr}
    \toprule
     MR (\%)     & HoMP  & HoRMP & HoOMP & GCG   & HaLRTC   & FP    & TMac  \\
     & & & & \cite{yu2014approximate} & \cite{liu2013tensor} & \cite{chen2014simultaneous} & \cite{xu2013parallel}\\
    \midrule
    50   & 16.09 & 20.95 & 45.02 & 126.14 & 42.89 & 88.18 & 23.00 \\
    60   & 13.85 & 17.77 & 37.38 & 106.60 & 54.38 & 97.64 & 26.56 \\
    70   & 11.41 & 14.27 & 28.29 & 85.72 & 72.21 & 121.94 & 32.75 \\
    80   & 8.63  & 10.66 & 19.84 & 64.13 & 98.76 & 149.37 & 42.78 \\
    90   & 5.25  & 6.28  & 10.79 & 38.89 & 97.03 & 100.28 & 49.41 \\
    95  & 3.45  & 3.94  & 6.05  & 24.69 & 96.72 & 71.72 & 46.44 \\
    99  & 1.98  & 2.05  & 2.35  & 14.41 & 1.98  & 54.65 & 46.06 \\

    \bottomrule
    \end{tabular}%
  \label{tab:results_tc_syn}%
\end{table}%

\begin{table*}[htbp]
\renewcommand{\arraystretch}{0.5}
  \centering
  \caption{Comparisons of different methods on tensor completion real datasets}
    \begin{mytabular}{rrrrrrrrrrrrrrrr}
    \toprule
          &       & \multicolumn{2}{c}{HoMP}          & \multicolumn{2}{c}{HoRMP}         & \multicolumn{2}{c}{HoOMP}        & \multicolumn{2}{c}{GCG \cite{yu2014approximate}}           & \multicolumn{2}{c}{HaLRTC \cite{liu2013tensor}}         & \multicolumn{2}{c}{FP \cite{chen2014simultaneous}}            & \multicolumn{2}{c}{TMac \cite{xu2013parallel}}     \\
       \cmidrule(r){3-4}   \cmidrule(r){5-6} \cmidrule(r){7-8}  \cmidrule(r){9-10}  \cmidrule(r){11-12}  \cmidrule(r){13-14}   \cmidrule(r){15-16}

    Datasets & MR (\%) & Relerr & Time (S.) & Relerr & Time (S.) & Relerr & Time (S.) & Relerr & Time (S.)  & Relerr & Time (S.) & Relerr & Time (S.) & Relerr & Time (S.)\\
    \toprule

          & 70    & 8.10E-02 & 1.17  & 8.16E-02 & 1.50  & 8.05E-02 & 2.64  & 7.85E-02 & 6.47  & 6.93E-02 & 32.09 & \underline{5.47E-02} & 195.12 & 2.45E-01 & 61.62 \\
    Lena  & 80    & 1.06E-01 & 0.92  & 1.07E-01 & 1.08  & 1.05E-01 & 1.78  & 9.70E-02 & 4.67  & 9.59E-02 & 44.14 & \underline{8.31E-02} & 199.84 & 4.01E-01 & 57.88 \\
    (512$\times$512$\times$3)      & 90    & 1.56E-01 & 0.31  & 1.56E-01 & 0.36  & 1.58E-01 & 0.42  & 1.42E-01 & 2.28  & 1.71E-01 & 68.94 & \underline{1.34E-01} & 64.42 & 7.15E-01 & 45.83 \\
          & 95    & 2.17E-01 & 0.24  & 2.19E-01 & 0.27  & 2.44E-01 & 0.32  & \underline{2.02E-01} & 1.66  & 3.77E-01 & 75.53 & 5.92E-01 & 32.04 & 9.09E-01 & 40.40 \\
          & 99    & 4.10E-01 & 0.21  & 4.38E-01 & 0.20  & 4.90E-01 & 0.22  & \underline{4.00E-01} & 1.05  & 8.61E-01 & 72.94 & 9.75E-01 & 14.03 & 1.01E+00 & 18.11 \\    
    \midrule
           & 70    & \underline{1.17E-02} & 43.37 & 1.18E-02 & 51.36 & \underline{1.17E-02} & 259.93 & 3.41E-02 & 47.62 & 3.00E-02 & 35.33 & 3.67E-02 & 442.16 & 1.81E-02 & 244.31 \\
     Spectral & 80    & 1.45E-02 & 34.76 & 1.45E-02 & 40.13 & \underline{1.44E-02} & 175.54 & 3.68E-02 & 35.24 & 4.24E-02 & 35.19 & 6.65E-02 & 444.89 & 2.65E-02 & 227.77 \\
     (205$\times$246$\times$96)      & 90    & 2.24E-02 & 24.40 & 2.23E-02 & 26.87 & \underline{2.20E-02} & 90.14 & 3.90E-02 & 21.97 & 6.46E-02 & 56.31 & 6.74E-02 & 210.08 & 6.31E-02 & 208.89 \\
           & 95    & 3.74E-02 & 7.34  & 3.80E-02 & 7.54  & \underline{3.68E-02} & 11.83 & 4.38E-02 & 14.78 & 1.05E-01 & 132.35 & 1.74E-01 & 91.23 & 1.71E-01 & 198.56 \\
           & 99    & 8.74E-02 & 2.25  & 8.97E-02 & 2.35  & 9.40E-02 & 2.47  & \underline{8.28E-02} & 9.59  & 8.50E-01 & 129.95 & 9.56E-01 & 33.66 & 6.94E-01 & 183.16 \\
     \midrule
 
          & 70    & 1.55E-02 & 42.00 & 1.54E-02 & 51.41 & 1.53E-02 & 331.32 & 5.98E-02 & 65.01 & \underline{3.22E-05} & 133.70 & 1.30E-03 & 86.05 & 5.81E-02 & 286.55 \\
    MRI   & 80    & 2.11E-02 & 33.20 & 2.14E-02 & 39.97 & 2.10E-02 & 177.95 & 6.54E-02 & 47.25 & 2.89E-03 & 148.18 & \underline{2.04E-03} & 116.64 & 9.68E-02 & 322.64 \\
   (181$\times$217$\times$181)       & 90    & 3.96E-02 & 21.44 & 3.97E-02 & 24.47 & 3.83E-02 & 89.69 & 7.32E-02 & 27.37 & 9.03E-02 & 158.40 & \underline{4.20E-03} & 150.70 & 2.26E-01 & 309.16 \\
          & 95    & 8.35E-02 & 7.13  & 8.30E-02 & 7.84  & \underline{7.95E-02} & 15.20 & 8.59E-02 & 16.90 & 3.44E-01 & 186.62 & 2.69E-01 & 120.94 & 4.47E-01 & 308.42 \\
          & 99    & \underline{3.18E-01} & 2.03  & 3.33E-01 & 2.24  & 3.46E-01 & 2.51  & 3.30E-01 & 9.17  & 9.95E-01 & 2.77  & 9.54E-01 & 52.91 & 9.08E-01 & 273.60 \\
    \midrule
 
          & 70    & 1.12E-01 & 90.95 & 1.14E-01 & 116.51 & 1.88E-01 & 57.08 & 1.79E-01 & 328.28 & 1.14E-01 & 288.78 & \underline{7.35E-02} & 697.61 & 1.89E-01 & 530.81 \\
    Knix1 & 80    & 1.26E-01 & 64.18 & 1.28E-01 & 80.70 & 2.03E-01 & 41.70 & 1.92E-01 & 281.14 & 1.60E-01 & 359.90 & \underline{7.85E-02} & 716.28 & 2.37E-01 & 904.77 \\
    (512$\times$512$\times$3$\times$22)      & 90    & 1.62E-01 & 37.69 & 1.62E-01 & 44.26 & 2.23E-01 & 23.14 & 2.09E-01 & 211.19 & 2.70E-01 & 381.45 & \underline{9.57E-02} & 714.48 & 3.25E-01 & 1237.13 \\
          & 95    & 2.21E-01 & 9.27  & 2.20E-01 & 10.51 & \underline{2.16E-01} & 38.80 & 2.24E-01 & 173.88 & 4.36E-01 & 383.07 & 6.05E-01 & 438.44 & 4.02E-01 & 1208.53 \\
          & 99    & 4.22E-01 & 2.26  & 4.14E-01 & 2.02  & 3.94E-01 & 5.92  & \underline{3.65E-01} & 134.75 & 9.27E-01 & 356.27 & 9.86E-01 & 138.46 & 8.26E-01 & 1098.88 \\
    \midrule
 
          & 70    & 1.34E-01 & 94.85 & 1.35E-01 & 121.01 & 1.93E-01 & 63.89 & 1.85E-01 & 337.07 & 1.50E-01 & 396.27 & \underline{8.05E-02} & 726.74 & 1.50E-01 & 562.98 \\
    Knix2 & 80    & 1.52E-01 & 66.01 & 1.54E-01 & 83.02 & 2.08E-01 & 41.80 & 2.00E-01 & 276.68 & 2.04E-01 & 390.28 & \underline{8.47E-02} & 732.19 & 1.92E-01 & 918.55 \\
    (512$\times$512$\times$3$\times$24)      & 90    & 1.99E-01 & 38.34 & 1.99E-01 & 45.50 & 2.28E-01 & 23.64 & 2.18E-01 & 208.10 & 2.98E-01 & 391.50 & \underline{1.19E-01} & 732.63 & 2.68E-01 & 1278.58 \\
          & 95    & 2.53E-01 & 9.57  & 2.53E-01 & 10.74 & \underline{2.42E-01} & 38.86 & 2.46E-01 & 167.50 & 3.89E-01 & 385.79 & 6.10E-01 & 441.75 & 3.86E-01 & 1230.64 \\
          & 99    & 3.85E-01 & 2.21  & 3.78E-01 & 2.00  & 3.80E-01 & 5.76  & \underline{3.52E-01} & 131.51 & 9.95E-01 & 7.80  & 9.85E-01 & 148.51 & 1.12E+00 & 1197.74 \\
          \midrule
 
     & 70    & \underline{7.11E-02} & 179.43 & 7.20E-02 & 211.71 & 1.07E-01 & 45.52 & 9.98E-02 & 360.14 & 7.42E-02 & 327.47 & 7.56E-02 & 1191.23 & 7.75E-02 & 1688.45 \\
    Tomato      & 80    & \underline{7.74E-02} & 133.91 & 7.77E-02 & 153.38 & 1.13E-01 & 29.72 & 1.07E-01 & 264.58 & 9.63E-02 & 436.98 & 7.98E-02 & 1203.06 & 8.33E-02 & 1597.08 \\
     (242$\times$320$\times$3$\times$167)     & 90    & \underline{8.51E-02} & 81.57 & 8.54E-02 & 87.78 & 1.21E-01 & 16.21 & 1.14E-01 & 158.06 & 1.35E-01 & 646.06 & 9.51E-02 & 1212.66 & 8.90E-02 & 1496.16 \\
          & 95    & \underline{9.55E-02} & 48.74 & \underline{9.55E-02} & 50.38 & 1.13E-01 & 25.52 & 1.17E-01 & 98.14 & 2.44E-01 & 634.48 & 4.32E-01 & 825.22 & 1.93E-01 & 1435.33 \\
          & 99    & 1.40E-01 & 14.01 & 1.40E-01 & 12.29 & 1.43E-01 & 28.90 & \underline{1.30E-01} & 41.69 & 8.81E-01 & 634.68 & 9.75E-01 & 293.10 & 7.41E-01 & 1389.03 \\          
    \bottomrule
    \end{mytabular}%
  \label{tab:tc_real}%
\end{table*}%

\subsubsection{Real data}
\begin{figure*}[htbp]
\centering

\renewcommand*{\arraystretch}{0.5}
\setlength{\tabcolsep}{1pt}

\begin{tabular}{cccccccc}

 \includegraphics[width=0.9in]
 {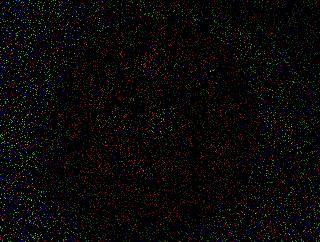}

&
 
 \includegraphics[width=0.9in]
 {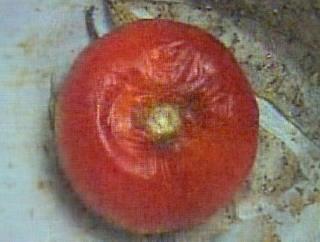}

&

 \includegraphics[width=0.9in]
 {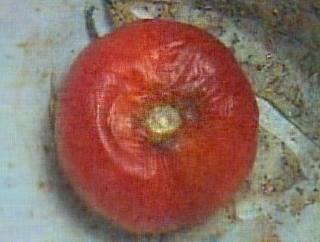}

&

 \includegraphics[width=0.9in]
 {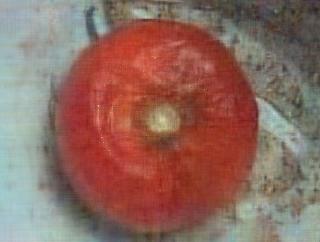}
 
 &
 
 \includegraphics[width=0.9in]
 {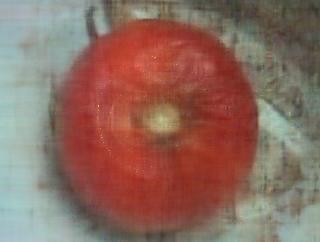}

 &
 
 \includegraphics[width=0.9in]
 {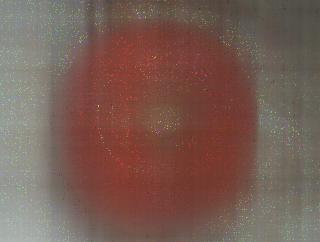}

&

 \includegraphics[width=0.9in]
 {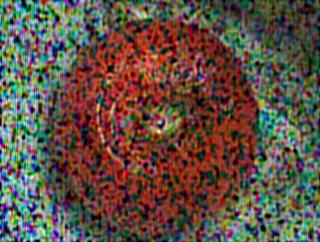}

&

 \includegraphics[width=0.9in]
 {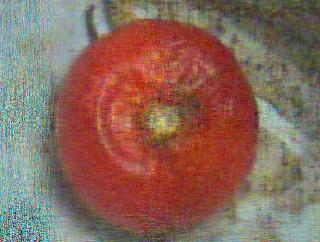}
 \\
 {\footnotesize (a) MR$=0.95$} & 
 {\footnotesize (b) HoMP} & {\footnotesize (c) HoRMP} & {\footnotesize HoOMP} & {\footnotesize (d) GCG \cite{yu2014approximate}} & {\footnotesize (e) HaLRTC \cite{liu2013tensor}} & {\footnotesize (f) FP \cite{chen2014simultaneous}} & {\footnotesize (g) TMac \cite{xu2013parallel}}

 \end{tabular}

 \caption{This example intuitively shows slides recovered by different methods from the 37-th slide of the Tomato dataset with MR$=0.95$
}\label{fig:tomato}
\end{figure*}

Several real data sets have been chosen: the Knix datasets, the Tomato video, the Hyperspectral images, the brain MRI and a color image      Lena\footnote{Knix    can be downloaded from \url{http://www.osirix-viewer.com/datasets/}, and Tomato, Hyperspectral images and brain MRI are available at \url{https://code.google.com/p/tensor-related-code/source/browse/trunk/Model/Tensor+Completion/LRTC_Package_Ji/?r=6}.}. Knix consists of two tensors of order-$4$,   Tomato is also a tensor of order-$4$, and the remaining datases are $3$rd order tensors. Since the order of magnitude of tensors in Knix and Tomato is $10^7$, for these datasets  we set the max iteration  of   FP as $50$, because it is too time-consuming at each iteration. For the same datasets  we also restrict the max iteration of HoOMP by $100$ because  at each iteration it has to solve a relatively large linear equation system. For HaLRTC on Tomato, the max iteration is set to $100$ due to efficiency.  The results are reported in Table \ref{tab:tc_real}, where we can observe that in most cases,  HoMPs have acceptable or better performances, but need   less time. HoMPs perform  particularly well on the Hyperspectral images: even the MR value is $99\%$, the relative error is less than $0.1$, which maybe due to the fact that the tensor  in this   dataset  is indeed low rank. On Tomato, which is of size $O(10^7)$, our methods take much less time than competing methods and obtain better performances.  On Knix and Lenna, FP outperforms other methods when the MR value is not high, which maybe because it uses some prior knowledges, however, mining the knowledges might be time-consuming, making it slower than other methods.
When recovering the color image, HoMPs have a   30x speedup comparing with HaLRTC, and   100x speedup comparing with FP, which may be useful in real world applications. In Fig. \ref{fig:tomato}, results recovered by different methods from one slide of the Tomato dataset with $0.95$ missing ratio are illustrated to intuitively evaluate the performances. From the results we can observe that HoMP and HoRMP recover more details than other methods.

\subsection{Multilinear multitask learning}

Our HoMP-type methods with least squares loss are compared with 4 state-of-the-art methods: sum of (overlapped) nuclear norm (Overlapped for short) \cite{romera2013multilinear,wimalawarne2014multitask} , latent nuclear norm (Latent) \cite{wimalawarne2014multitask} , scaled latent nuclear norm (Scaled) \cite{wimalawarne2014multitask}, and a nonconvex approach (Nonconvex) \cite{romera2013multilinear}. The first three methods are based on nuclear norm regularized convex optimization, while the last method factorizes the tensor into factors, which is nonconvex. We also note here that the difference between the Overlapped method in \cite{romera2013multilinear} and \cite{wimalawarne2014multitask} is that \cite{romera2013multilinear} uses the sum of nuclear norm as a regularization, while \cite{wimalawarne2014multitask} treats it as a constraint. In the following we consider them as the same method. The stopping criterion is the same as the previous experiment.   Parameters are tuned via $10$-fold cross validation. Specifically, $K\in\{15,20,25,\ldots,50 \}$. The following datasets are chosen:

1) School dataset. This datasets is  made available by the Inner London Education Authority (ILEA), which consists of examination records from $139$  schools in years $1985$, $1986$ and $1987$, with $15362$ students. Each task is to predict exam scores for students in each school, where the size of the input space is $25$, with one indicating the bias term. Following \cite{wimalawarne2014multitask}, we model it as a MLMTL problem, where each task has two indices: the school index and the year index. Therefore, the tasks jointly gives a $25\times 3\times 139$ weight tensor to be learned. The size of the training set varies from $2000$ to $12000$. Similar to \cite{wimalawarne2014multitask}, we use the explained variance $100\cdot(1- {\rm MSE_{test}}/{\rm var}(\mathbf y))  $ to evaluate the performance of the compared methods.

2) Restaurant \& consumer dataset \cite{vargas2011effects}\footnote{We would like to thank the first author of \cite{romera2013multilinear} to provide us the cleaned dataset.}.  This dataset consists of rating scores from $138$ consumers to different restaurant from $3$ aspects, with $3483$ instances. Each task is to predict the rating given by a consumer from one aspect, provided a restaurant as an input. The size of the input space is $45$, with one indicating the bias term. Since each task can be indexed by two indices: the consumer index and the aspect index, the tasks jointly yields a $45\times3\times138$ weight tensor to be learned. The size of the training set varies from $400$ to $2000$. The test MSE is used to evaluate the performance of the compared methods.

\begin{figure}[H] 
   \centering
      \begin{subfigure}[b]{0.45\textwidth}
                 \includegraphics[width=\textwidth]{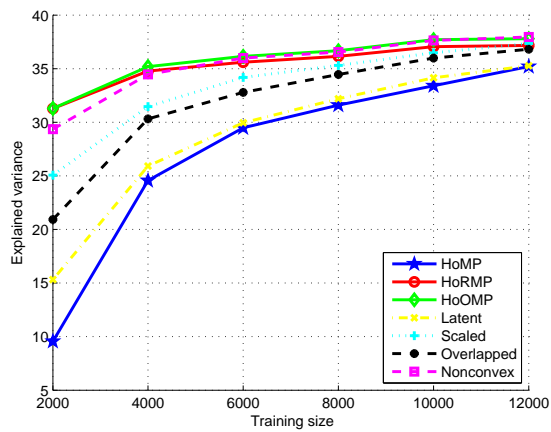}
                  \caption{Explained variance (the larger, the better) for different methods on the school dataset. }  \label{fig:mlmtl_school}          \end{subfigure}   \\
      \begin{subfigure}[b]{0.45\textwidth}
                 \includegraphics[width=\textwidth]{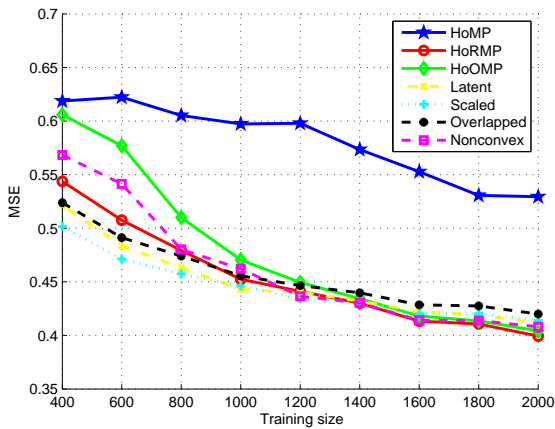}
    \caption{Test MSE for different methods on the restaurant dataset.} \label{fig:mlmtl_res}  \end{subfigure} 
    \begin{subfigure}[b]{0.45\textwidth}
      \includegraphics[width=\textwidth]{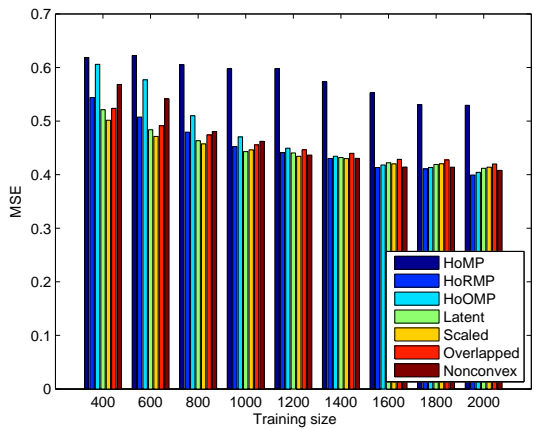}
   \caption{Test MSE   on the restaurant data  using bar plot.} \label{fig:mlmtl_res_bar} \end{subfigure}                                     
      \caption{MLMTL results on two real datasets with different methods: HoMP, HoRMP, HoOMP, Latent \cite{wimalawarne2014multitask}, Scaled \cite{wimalawarne2014multitask}, Overlapped \cite{romera2013multilinear,wimalawarne2014multitask} and Nonconvex  \cite{romera2013multilinear}. }  \label{fig:mlmtl}               
\end{figure}

The results averaged over $20$ instances on the school and the restaurant datasets are respectively reported in Fig. \ref{fig:mlmtl_school} and \ref{fig:mlmtl_res}. From Fig. \ref{fig:mlmtl_school}, we can observe that HoRMP and HoOMP perform comparable with or better than other methods on the school dataset, while HoMP does not perform well. Scaled \cite{wimalawarne2014multitask} performs best among Latent \cite{wimalawarne2014multitask}, Scaled \cite{wimalawarne2014multitask} and Overlapped \cite{romera2013multilinear,wimalawarne2014multitask}, which is in accordance with the observations in \cite{wimalawarne2014multitask}. The method Nonconvex \cite{romera2013multilinear} is slightly worse than HoRMP and HoOMP when the size of the sample is small, and then it catches up our methods as the samples increase. For the restaurant dataset, HoMP is still worse than other methods, see Fig. \ref{fig:mlmtl_res}; HoRMP and HoOMP are not as good as other methods when the sample size is small, which may be due to the overfitting in these cases. However, they eventually catch up other methods when the samples increase, and are slightly better  when the sample size is larger than $1600$.    To give a clearer view on their performances, we also use the bar plot to show the MSE of the compared methods in Fig. \ref{fig:mlmtl_res_bar}. The efficiency comparisons are reported in Table \ref{tab:mlmtl_time}, from which one can again see the efficiency of HoMP-type methods.

During the experiments, we also observe that for our methods, around $K=25$ iterations can give desirable results, which implies that the weight tensor $\mathcal W$ can indeed be approximated by a tensor of rank lower than $25$.

\begin{table} 
\renewcommand{\arraystretch}{0.5}
  \centering
  \caption{Efficiency comparison of different methods on school and restaurant datasets. The results are averaged over all sample sizes and all the instances.}\label{tab:mlmtl_time}
    \begin{mytabular1}{rrrrrrrr}
    \toprule
    Dataset   & HoMP  & HoRMP & HoOMP & Latent \cite{wimalawarne2014multitask}& Scaled \cite{wimalawarne2014multitask}& Overlapped \cite{romera2013multilinear,wimalawarne2014multitask}& Nonconvex \cite{romera2013multilinear}\\
    \midrule
    School & 0.33  & 0.35  & 0.36  & 4.92  & 1.44  & 8.55  & 110.31 \\
    Restaurant & 0.45  & 0.46  & 0.47  & 1.35  & 5.81  & 8.37  & 110.17 \\
    \bottomrule
    \end{mytabular1}%
  \label{tab:addlabel}%
\end{table}%

\subsection{Robust tensor completion}
The goal of this subsection is to examine the effectiveness of the HoMP-G strategy. A suitable setting is the robust tensor completion, with the loss function   instantiated by the Cauchy loss $\ell_{\sigma}(t) = \sigma^2/2\log( 1+ t^2/\sigma^2)$ with $\sigma$ fixed to $0.08$. The compared methods are HoRPCA \cite[Alg. 2.2]{goldfarb2014robust} and RPCA \cite{candes2011robust}. HoRPCA and RPCA are designed to solve convex optimization problems, which employ the $\ell_1$ loss to penalize the noise or outliers, where RPCA is focused on robust matrix completion, while HoRPCA is for   robust tensor completion. 
The stopping criterion and other settings are the same as the previous experiments. 

\subsubsection{Synthetic data}
\begin{figure}[H]
   \centering
      \begin{subfigure}[b]{0.5\textwidth}
                 \includegraphics[width=\textwidth]{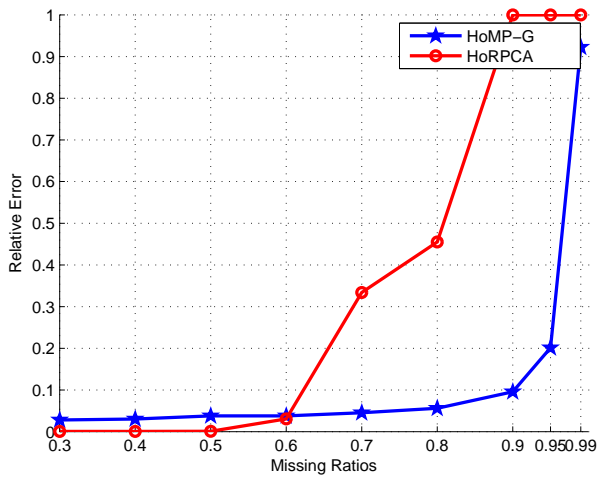}
          \end{subfigure} 
                   \caption{Robust tensor completion results on synthetic data ($100\times100\times100$,   CP-rank $=10$)  in terms of relative error. The compared method is HoRPCA \cite{goldfarb2014robust}.}    \label{fig:rtc_syn}
\end{figure}

Third order tensors of size  $100\times100\times100$ are randomly generated, with CP-rank $10$. Some entries are randomly missing, with missing ratio (MR) varies from $0.3$ to $0.99$. $10\%$ of the entries are contaminated by outliers drawn  form $[-1,1]$.  We compare with HoRPCA in this experiment, and report the results in Fig. \ref{fig:rtc_syn}. We can observe that when the MR value is less than $0.6$, HoRPCA is better than our method; when the MR value increases, the performance of HoRPCA decreases rapidly, while our method is more stable. This observation is similar to that in the experiment of tensor completion.

\subsubsection{Yale face} 
\begin{figure}[htbp]
\centering

\renewcommand*{\arraystretch}{0.5}
\setlength{\tabcolsep}{1pt}

\begin{tabular}{ccccc}
 
 (a)
 &

 \includegraphics[width=0.756in,height=0.864in]
 {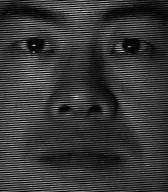}

&

 \includegraphics[width=0.756in,height=0.864in]
 {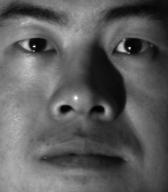}

&

 \includegraphics[width=0.756in,height=0.864in]
 {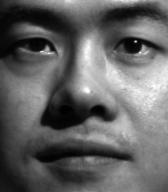}
 
 &
 
 \includegraphics[width=0.756in,height=0.864in]
 {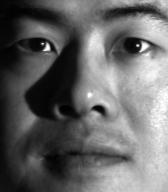} 
 
\\
 
 (b)
 &
 
 \includegraphics[width=0.756in,height=0.864in]
 {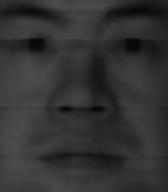}

&

 \includegraphics[width=0.756in,height=0.864in]
 {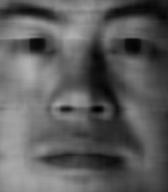}

&

 \includegraphics[width=0.756in,height=0.864in]
 {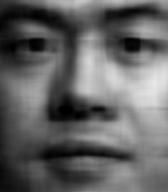}
 
 &
 
 \includegraphics[width=0.756in,height=0.864in]
 {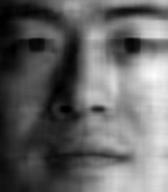} 
 
\\ 

 (c)
 &
 
 \includegraphics[width=0.756in,height=0.864in]
 {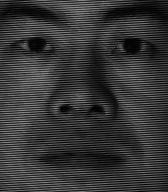}

&

 \includegraphics[width=0.756in,height=0.864in]
 {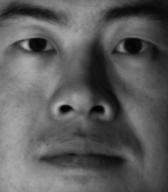}

&

 \includegraphics[width=0.756in,height=0.864in]
 {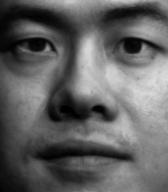}
 
 &
 
 \includegraphics[width=0.756in,height=0.864in]
 {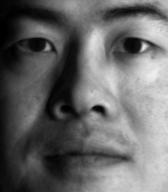} 
 
\\ 

 (d)
 &
 
 \includegraphics[width=0.756in,height=0.864in]
 {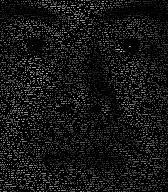}

&

 \includegraphics[width=0.756in,height=0.864in]
 {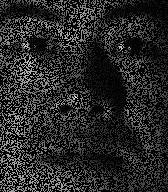}

&

 \includegraphics[width=0.756in,height=0.864in]
 {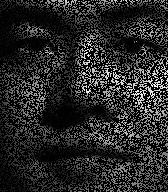}
 
 &
 
 \includegraphics[width=0.756in,height=0.864in]
 {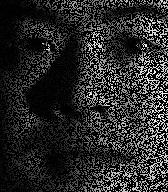} 
 
\\

 (e)
 &
 
 \includegraphics[width=0.756in,height=0.864in]
 {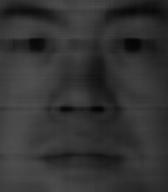}

&

 \includegraphics[width=0.756in,height=0.864in]
 {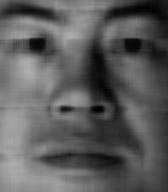}

&

 \includegraphics[width=0.756in,height=0.864in]
 {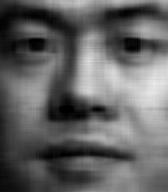}
 
 &
 
 \includegraphics[width=0.756in,height=0.864in]
 {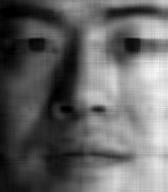} 
 
\\ 

 (f)
 &

 \includegraphics[width=0.756in,height=0.864in]
 {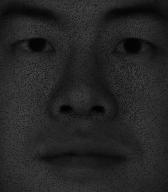}

&

 \includegraphics[width=0.756in,height=0.864in]
 {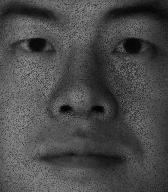}

&

 \includegraphics[width=0.756in,height=0.864in]
 {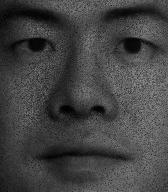}
 
 &
 
 \includegraphics[width=0.756in,height=0.864in]
 {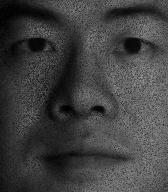} 
 
\\ 
 
 \end{tabular}

 \caption{Comparison of HoMP-G and RPCA on removing shadows and specularities from face images. (a): Some original images. (b) Images recovered by HoMP-G. (c) Images recovered by RPCA. (d) MR $=60\%$. (e) Recovered by HoMP-G. (f) Recovered by RPCA. 
}\label{fig:rtc_face}
\end{figure}

We compare our method and RPCA on removing shadows and specularities from face images, as that done in \cite{candes2011robust}. The shadows and specularities can be seen as noise or outliers \cite{candes2011robust}. The selected dataset consists of $64$ face images of size $192\times 168$, giving a $192\times168\times64$ tensor. Previously RPCA treats it as a $32256\times 64$ matrix \cite{candes2011robust}, while we directly treat it as a tensor. We consider two settings: there are no missing entries and there are $60\%$ missing entries. The maximum iteration for both  two methods is $200$. Part of the results are shown in Fig. \ref{fig:rtc_face}. In the figure,  row (a) shows  some original images; (b) presents  images recovered by HoMP-G (HoMP for short in this paragraph) and (c) are those recovered by RPCA. We can observe that both of the two methods can remove the shadows, while it seems from the first column that HoMP performs better, as it can remove the lines. However, images recovered by HoMP are not as clear as those recovered by RPCA. This may be because that HoMP yields a tensor of CP-rank $200$, say $\sum^{200}_{i=1}\alpha^i\mathbf x_i\otimes \mathbf y_i\otimes\mathbf z_i$, to approximate the original data tensor, which is equal to that for the $j$-th image, $1\leq j\leq 64$, it is approximated by a matrix of rank at most $200$, $\sum^{200}_{i=1}\alpha^i z_{i,j}\mathbf x_i\otimes\mathbf y_i$. Since these $\mathbf x_i$ and $\mathbf y_i$ are not orthogonal, they may not be principle components, and hence may lead to loss of information. Rows (d)--(f) show the images with $60\%$ entries missing and those recovered by HoMP and RPCA, respectively. In this case HoMP still performs stable, while RPCA cannot successfully impute all the missing entries. Finally, HoMP only requires $(192+168+64)\cdot 200 = 169600$ size to store the recovered tensor, while RPCA  has to use $2064384$ size to store the recovered matrix. Totally speaking, our method has the following features: it can remove shadows and specularities, impute missing entires, as well as generate  a set of common basis $\{ \mathbf x_i,\mathbf y_i \}$ for all the images, which has lower storage requirement, and   is as efficient as the     previous HoMPs.
 
\subsection{Linear convergence}
\begin{figure} 
   \centering
      \begin{subfigure}[b]{0.4\textwidth}
                 \includegraphics[width=\textwidth]{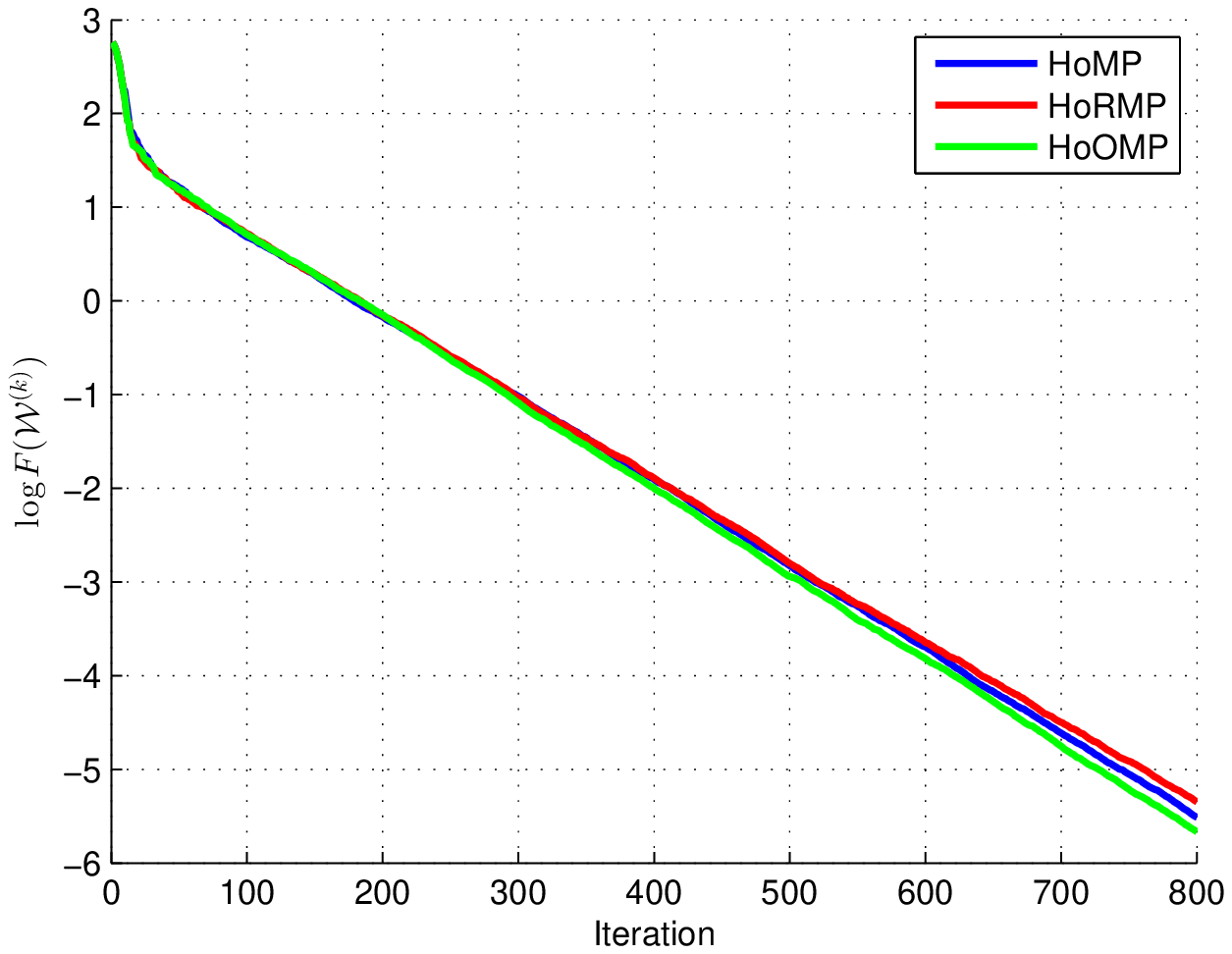}\caption{Convergence rate on tensor completion}\label{fig:lc_tc}
          \end{subfigure} 
      \begin{subfigure}[b]{0.4\textwidth}
                 \includegraphics[width=\textwidth]{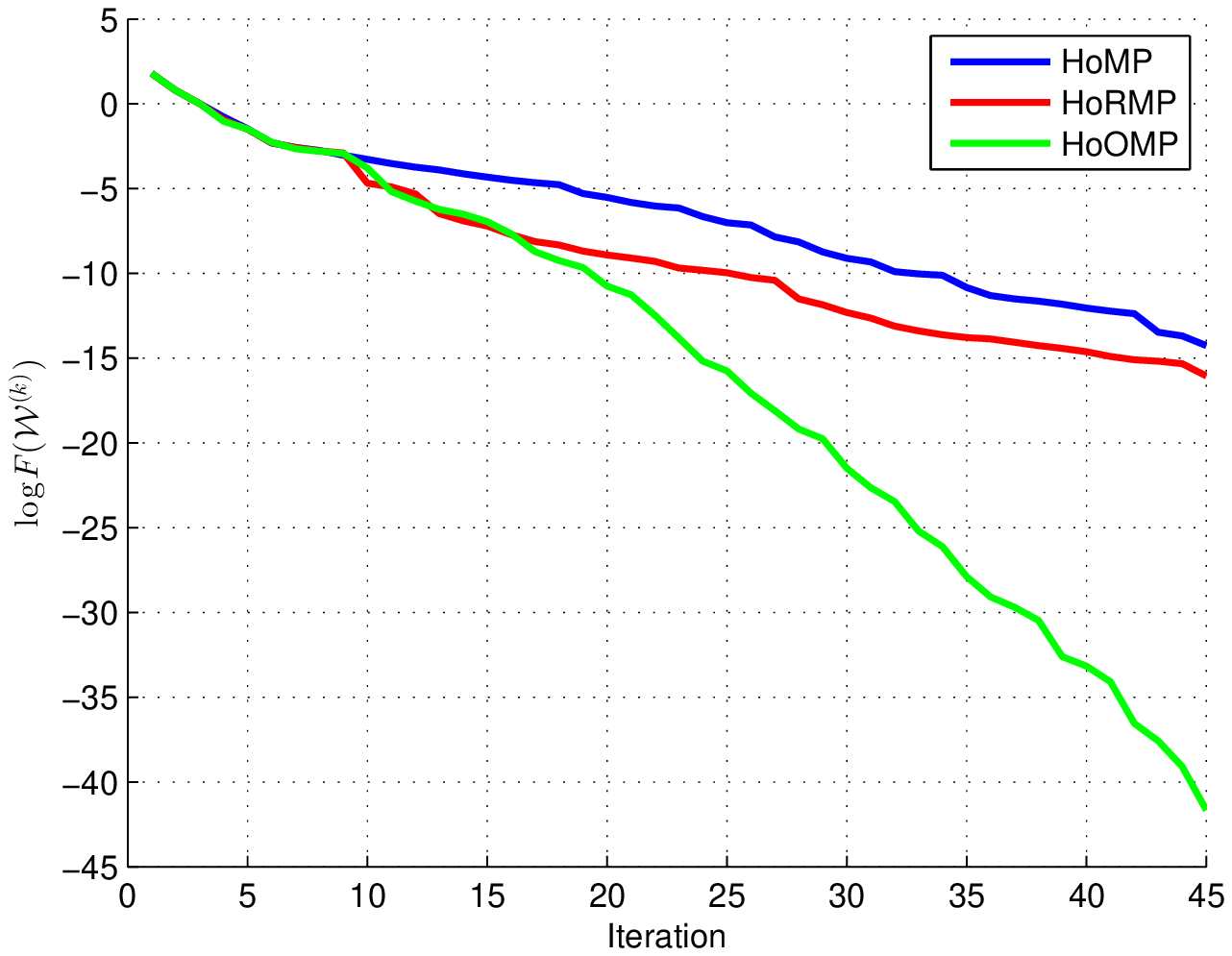}\caption{Convergence rate on MLMTL}\label{fig:lc_mlmtl}
          \end{subfigure}    
      \begin{subfigure}[b]{0.4\textwidth}
                 \includegraphics[width=\textwidth]{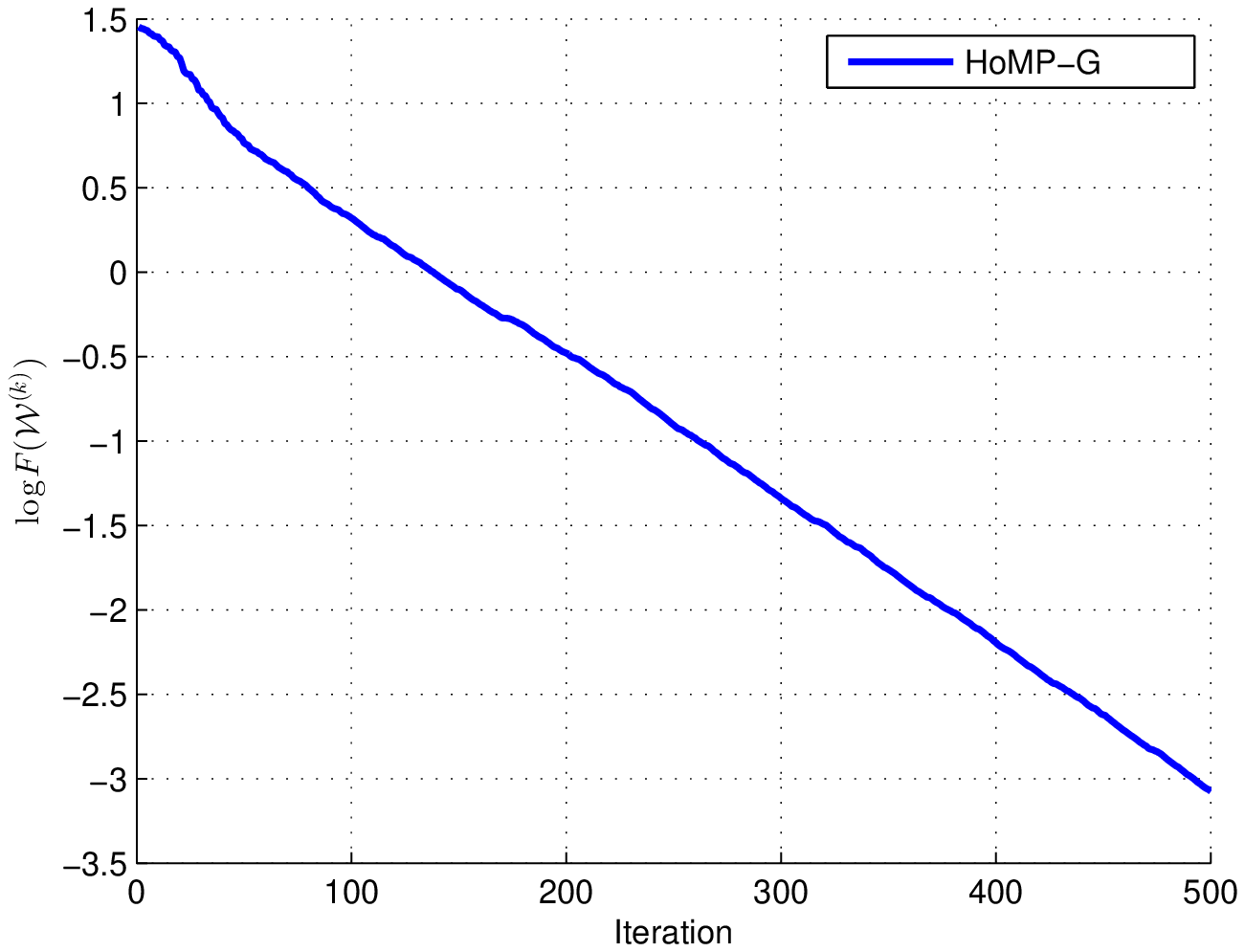}\caption{Convergence rate on robust tensor completion}\label{fig:lc_rtc}
          \end{subfigure}                
                   \caption{Convergence rate of HoMPs on different experiments. Fig. \ref{fig:lc_tc}: tensor completion. Fig.  \ref{fig:lc_mlmtl}: MLMTL. Fig. \ref{fig:lc_rtc}: robust tensor completion.  $y$-axes:  the logarithm of the cost function at iteration $k$, $\log( F(\mathcal W^{(k)}))$; $x$-axes:iteration. The plots confirms the theoretical results in Sect. \ref{sec:lc}.}    \label{fig:linear_convergence}
\end{figure}

Last, we examine the linear convergence of HoMPs on three experiments, as shown in Fig. \ref{fig:linear_convergence}. In the figures, the $y$-axes is the logarithm of the cost function at iteration $k$, $\log( F(\mathcal W^{(k)}))$, while $x$-axes stands for iteration. Specifically, Fig. \ref{fig:lc_tc} plots the curves of HoMP-LS (blue), HoRMP-LS (red) and HoOMP-LS (green) on   tensor completion; Fig. \ref{fig:lc_mlmtl} plots those on MLMTL, while Fig. \ref{fig:lc_rtc} plots the curve of HoMP-G on robust tensor completion. From the figures, we can observe that the curves confirm the theoretical results derived in Sect. \ref{sec:lc}.  

 \section{Conclusion}\label{sec:con}
In this paper, we proposed higher order matching pursuit for low rank tensor learning. Comparing with some state-of-the-art methods, HoMPs have three important features: low computational complexity, low storage requirement, and linear convergence. Furthermore, HoMP can also be applied to problems with a nonconvex cost function, sharing the same convergence rate as those with a convex cost function. Numerical experiments on synthetic as well as real datasets verify the efficiency and effectiveness of HoMPs. 
 
 \section*{Acknowledgement}{ \scriptsize
  The research leading to these results has received funding from the European Research Council under the European Union's Seventh Framework Programme (FP7/2007-2013) / ERC AdG A-DATADRIVE-B (290923). This paper reflects only the authors' views, the Union is not liable for any use that may be made of the contained information;
  Research Council KUL: GOA/10/09 MaNet, CoE PFV/10/002 (OPTEC),
  BIL12/11T; PhD/Postdoc grants; Flemish
    Government: FWO: PhD/Postdoc grants, projects: G.0377.12 (Structured systems), G.088114N (Tensor based	
    data similarity); 
    IWT: PhD/Postdoc grants, projects: SBO POM (100031); 
    iMinds Medical Information Technologies SBO 2014; 
    Belgian Federal Science Policy Office: IUAP P7/19 (DYSCO, Dynamical
    systems, control and optimization, 2012-2017). Johan Suykens is a professor at KU Leuven, Belgium.}
 
  \bibliographystyle{IEEEtran}      
 \bibliography{TensorCompletion,rank-1_tensor,cg_fw,multitask,MatchingPursuit}
\end{document}